\documentclass[a4paper, 9pt]{article}
\usepackage[utf8]{inputenc}

\usepackage[9pt]{extsizes}
\usepackage[left=30mm, right=15mm, top=20mm, bottom=20mm, bindingoffset=0cm]{geometry}

\usepackage{amsfonts}

\usepackage{graphicx}
\usepackage{wrapfig}

\usepackage{caption}

\usepackage{bigstrut}
\usepackage{multirow}
\usepackage{array}
\usepackage{tabularx}

\usepackage[T2A]{fontenc}
\usepackage[utf8]{inputenc}
\usepackage{hyperref}
\usepackage[rgb]{xcolor}
\hypersetup{
    colorlinks=true,       	% false: ссылки в рамках
	urlcolor=blue          % на URL
	}
	
\usepackage[symbol*]{footmisc}

\usepackage{setspace}

\usepackage{amssymb}

\begin{document}

%\maketitle

\begin{center}
\begin{spacing}{1.8}

\textbf{\LARGE{Human Activity Recognition using Continuous Wavelet Transform }}

\textbf{\LARGE{and Convolutional Neural Networks}}

\LARGE{\mbox{Anna Nedorubova \footnotemark}\footnotetext{nedorubova.aa@phystech.edu},
Alena Kadyrova, 
\mbox{Aleksey Khlyupin\footnotemark}\footnotetext{khlyupin@phystech.edu}}

\Large{\textit{Center for Engineering and Technology of MIPT,}}

\Large{\textit{Moscow Institute of Physics and Technology,}}

\Large{\textit{Institutskiy Pereulok 9, Dolgoprudny, Moscow 141700, Russia}}

\vspace{100px}


\end{spacing}

\end{center}

\newpage

\setcounter{tocdepth}{2}
{
  \hypersetup{linkcolor=black}
  \tableofcontents
}

\newpage

\Large{\textbf{Abstract}}

Quite a few people in the world have to stay under permanent surveillance for health reasons; they include diabetic people or people with some other chronic conditions, the elderly and the disabled. These groups may face heightened risk of having life-threatening falls or of being struck by a syncope. Due to limited availability of resources a substantial part of people at risk can not receive necessary monitoring and thus are exposed to excessive danger. Nowadays, this problem is usually solved via applying Human Activity Recognition (HAR) methods.
%Сontinuous monitoring of the elderly and the disabled is an important social concern, however, not all of them can afford being under constant supervision of nurses, sitters or relatives. This problem is solved via applying Human Activity Recognition (HAR) methods. 
HAR is a perspective and fast-paced Data Science field, which has a wide range of application areas such as healthcare, sport, security, surveillance etc. 
However, the currently techniques of recognition are markedly lacking in accuracy, hence, the present paper suggests a highly accurate method for human activity classification.
We propose a new workflow to address the HAR problem and evaluate it on the UniMiB SHAR dataset \cite{UniMiB}, which consists of the accelerometer signals. The model we suggest is based on continuous wavelet transform (CWT) and convolutional neural networks (CNNs). 
%Wavelet transform is a variation of the Generalized Fourier transform which uses aperiodiс wavelets instead of the periodic harmonic functions. 
Wavelet transform localizes signal features both in time and frequency domains and after that a CNN extracts these features and recognizes activity. It is also worth noting that CWT converts 1D accelerometer signal into 2D images and thus enables to obtain better results as 2D networks have a significantly higher predictive capacity. In the course of the work we build a convolutional neural network and vary such model parameters as number of spatial axes, number of convolutional and dense layers, number of neurons in each layer, image size, type of mother wavelet, the order of zero moment of mother wavelet etc. Besides, we also apply models with residual blocks which resulted in significantly higher metric values. Finally, we succeed to reach 99.26 \% accuracy and it is a worthy performance for this problem.

\section{Introduction}

%Quite a few people in the world have to stay under permanent surveillance for health reasons; they include diabetic people or people with some other chronic conditions, the elderly and the disabled. These groups may face heightened risk of having life-threatening falls or of being struck by a syncope. Due to limited availability of resources a substantial part of people at risk can not receive necessary monitoring and thus are exposed to excessive danger. Nowadays, this problem is usually solved via HAR-based supervision systems. However, the currently techniques of recognition are markedly lacking in accuracy. Hence, the present paper suggests a highly accurate method for human activity classification which can be used for surveillance systems beneficial for elderly, disabled and diseased.

Human Activity Recognition (or HAR) \cite{har} is a promising and rapidly growing branch of Data Science, which nowadays have scores of directions in application. Firstly, HAR is widely used in the area of smart homes \cite{home}, where it plays two significant roles: health care of the elderly and disabled and adaptation of the environment to the residents' habits in order to improve the quality of their lives \cite{cognitive}. Then, another area of HAR implementing is the security and surveillance sphere\cite{areas}. For example, in airports surveillence systems can detect about 50 different types of actions such as aircraft arrival preparation or baggage unloading, and in public places like metro special cameras can recognize fighting and vandalism. Besides, HAR is often applied for different sport purposes: not only by individuals for their private objectives (such as keeping the history of their sport activity, struggling with the Office Workers Syndrome \cite{syndrom} of calculating daily energy loss), but also in the field of great sport for event detection for highlight generation and automatic video commenting \cite{sport}.

Finally, as it has already been mentioned, one of the most popular and, at the same time, socially beneficial area of HAR application is healthcare. Elderly people or people vulnerable to syncope (such as diabetics) have to be kept under continuous observation, what sometimes can be rather difficult. One method of combating this problem is to record any particular signal connected with person movements and to process it  automatically. If any data anomaly is detected, it will be a cause for concern that something like fall or syncope has occurred.

%\cite{har}

\begin{figure}[h]
    \centering
    \includegraphics[scale=0.15]{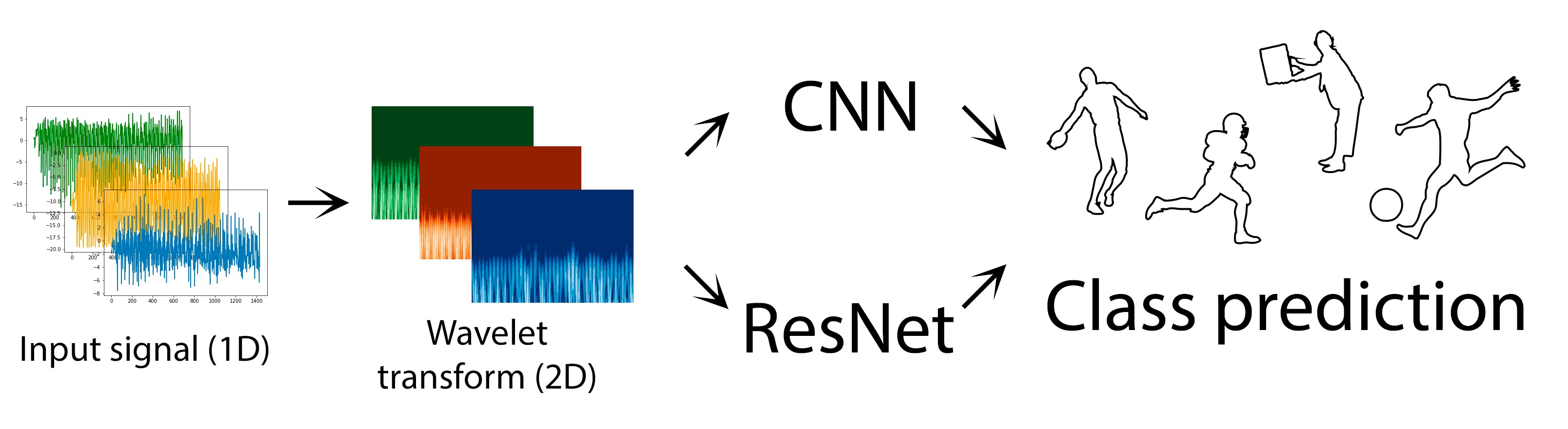}
    \caption{Workflow consists of several parts. Firstly, 1D accelerometer signals are converted into 2D images via wavelet transform. Then, a neural network (CNN or ResNet) is trained on these images and, finally, it becomes capable of predicting the class of human activity}
    \label{pic1_TOC}
\end{figure}

Generally speaking, the process of HAR can be roughly divided into two parts - data collection and data processing. The two main types of data-collecting devices are exterior and wearable sensors. \cite{wear} Each of these types has its benefits and drawbacks. Intelligent homes can be a classic illustration of exterior detection systems. The dispersion of sensor placement provides a better opportunity for data collection and, thereafter, for the activity identification. Separate exterior cameras are also frequently used for detection and recognition. They can simultaneously follow a few people and they do not burden observed individuals with their weight. However, exterior sensors have a number of shortcomings. Firstly, external devices should be connected with a certain place and has a limited angle of aspect. Then, video recording can be produced hiddenly and it is illegally. Finally, video processing requires much more resources compared with the processing of some other types of signals, which will be discussed further.

Wearable sensors is an excellent alternative for external detectors. There is plenty of parameters that can be measured by them for activity detection. For example, outside condition can indicate the type of activity: if it is noiseless and dark, the object is probably sleeping, if it is cold, the object may walking outside, etc. Physiological signals (heart and breath rate, skin temperature, blood pressure etc) also carry some information connected with activity. Finally, there is a wide range of variable physical quantities which can be used as an input signal. For measuring this signal corresponding apparata are used: accelerometers, gravimeters, magnetometers and dozens more.  It is noteworthy that these apparata are often built in smartphones and can be managed just by means of a mobile app.  Therefore, wearable sensors turned out to be the most suitable type of device to meet our target. In the course of the current project we use a dataset UniMiB-SHAR \cite{UniMiB}, which contains benchmarked accelerometer data.

Thus, our objective is to determine the type of human activity based on the accelerometer signal \cite{daily_active}. There are numbers of classifies, which can be appropriate for this classification problem; following are some of them. 

Firstly, there is \textit{Naive Bayes}, which assumes that all the features are independent and classifies every object with the probability proportional to conditional probabilities of its features \cite{naive}. In comparison to other methods, Naive Bayes performs a markedly lower accuracy \cite{bayes_ex}.
%Naive Bayes, assumes that all the features are independent and classifies every object with the probability proportional to conditional probabilities of its features \cite{naive}. Based on the Bayes rule, the posterior probability is calculated and so the class label is determined. The main disadvantage of this method is that all the features have to be independent, otherwise the Bayes rule is unjust. However, Naive Bayes is not is not very frequently used for activity classification because of its relatively low classification capacity. For example, it can be seen from the article \textit{Smoothed Naive Bayes-Based Classifier for Activity Recognition}  \cite{bayes_ex} that compared to other methods, Naive Bayes performs a markedly lower accuracy.
Another classifier, \textit{Support Vector Machine} (or SVM) divides objects in such a way as to maximize a gap between classes and, if objects are not separable linearly, increases the dimension of space \cite{svm}. The authors of the UniMiB SHAR dataset managed to gain 78.75\% accuracy via the SVM method \cite{UniMiB}. At the same time, \textit{Random Forest} (a group of \textit{Decision Trees}\cite{tree}, \cite{forest}) enabled authors to obtain the 81.48\% accuracy.
%Another classifier, Support Vector Machine (or SVM) divides objects in such a way as to maximize a gap between classes and, if objects are not separable linearly, increases the dimension of space \cite{svm}. The main principle of this method is to separate the set of data with a hyper-plane the way when objects are as far from it as possible. When no linear separation can be achieved, special 'kernels' allow to move to the non-linear feature space. The drafters of the UniMiB SHAR dataset gained 78.75\% accuracy via the SVM method. Their work is presented in the article \textit{UniMiB SHAR: A Dataset for Human Activity Recognition Using Acceleration Data from Smartphones} \cite{UniMiB} (если я уже ссылалась, то надо ли еще раз?).
%Decision Tree is a recursive classifier, consisted of edges and leaves. \cite{tree}. Edges have objective functions and leaves have objective function values. In the process of classification an objective value descents to an appropriate leaf and thus the process of classification is being executed. A group of decision trees is intuitively called Random Forest (RF). Every tree independently makes a prediction. The decision of the forest is taken subject to all the tree "votes" \cite{forest}. Random Forest was also applied by the authors of the article \textit{UniMiB SHAR: A Dataset for Human Activity Recognition Using Acceleration Data from Smartphones}. In this case they obtained 81.48\% accuracy.
\textit{K Nearest Neighbours} (kNN) classifier ranks an object with the group which the majority of its neighbours belongs to \cite{knn}. This method is applied in the article \cite{knn_ex}, where quite high results, compared to some other methods such as Decision Trees, are presented.
%The authors developed a model with k=5 and managed to obtain 92.32\% accuracy. 
%K Nearest Neighbors (kNN) classifier ranks an object with the group which the majority of its neighbors belongs to \cite{knn}. In fact, every step consists of two actions: firstly, k nearest neighbours are determined and then based on them the class label is assigned. Sometimes, the nearer neighbour is the more weight it has while class determination. K Nearest Neighbour classifier requires that all the data should be normalized. This method is described in the article  \textit{Activities of daily living and falls recognition and classification from the wearable sensors data} \cite{knn_ex}, for example. According to this article, the kNN method presents quite high results compared to some other methods such as Desicion Trees.
%The authors developed a model with k=5 and managed to obtain 92.32\% accuracy. 
\textit{Logistic Regression} (LR) considers class label as a probability function and maximize the likelihood function \cite{regr}. This method is designed for binary classification and can be used in the HAR purpose only in particular cases \cite{LR}. However, as the authors \cite{LR} notice, sometimes LR is significantly more time-consuming than some of the algorithms and as time-consuming as neural networks whereas it occasionally presents low results compared with other techniques.

Finally, different types of \textit{Artificial Neural Networks} (ANNs) are widely used for classification problems \cite{ann}, \cite{yeg}. Due to their layered structure and a high number of weights, ANNs are able to capture even implicit features, thereby being an excellent classifier. The layers of the ANNs are composed of neurons, which are essentially primitive computational processors. Neural Networks differ in the architecture; in the current work \textit{the Convolutional Neural Networks} (or CNNs) \cite{four}, \cite{wu} are discussed.

In addition to dense layers, which are contained in almost all types of networks, CNNs also include convolutional and pooling layers. Convolutional layers enable to extract image features into feature maps, pooling layers decreases feature map dimension.

%A typical CNN is composed of three types of layers: dense, convolutional  and pooling.  

%The most specific feature of CNNs is a convolution function: a small convolution matrix runs through a layer and elementwise multiplies with its elements. The resulting matrix is called a feature map. The pooling layers are used to decrease the feature maps dimension. CNNs were invented by analogy with visual cortex. Convolutional layers allow to distinguish major features and neglect minor and are insensitive to feature size, coordinates and orientation. All of this make CNNs a perfect solution for computer vision and image classification problems. However, so-called vanishing gradient can occur in deep CNN layers sometimes. This problem results in weight immutability, and thus the training process stops. Residual blocks is what make it possible to  overstep this obstacle \cite{residual}. A residual block adds a shortcut connection between a pair of layers. The function of residual block is as follows:

%The addition $\mathcal{F}(x, W_i)+x$ is precisely what a shortcut connection does.

A perfect example of ANNs application for human activity recognition is described by Debadyuti Mukherjee, et al. in the article \cite{ensemble}. Here the authors suggest a model composed of three independent neural networks: CNN-Net,  Encoded-Net and  CNN-LSTM. Each of this models has a special role. Namely, the CNN-Net model is in charge of feature extraction, the Encoded-Net model searches for hidden information and represents it in a simpler way, while the CNN-LSTM model keeps information fluctuations according to its sequence. The combined model, called EnsemConvNet, enabled to obtain 92.6\% accuracy on the UniMiB SHAR dataset.

Another CNN-based method is presented in the article \cite{modified} by Alireza Keshavarzian, et al. Their model consists of three parts. The first part is composed of one-dimensional convolution layer, batch normalization and ReLU activation function. In parallel with it the second part is situated, which is a pile of five residual blocks. Lastly, the architecture is topped with a module which bring these two branches together and gives the final answer. This approach enabled to reach 93.4 \% accuracy on the UniMiB SHAR dataset. 

In some cases CNNs are used in combination not only with other networks but also with different methods. For example, in the article \cite{ronao} it is shown that extracting features via Fourier transform can significantly increase accuracy up to 94.02\%.

%Along with developing residual networks of their own, researchers frequently use off-the-shelf models. Some of the Python libraries, namely PyTorch, TensorFlow and Keras, provide ready solutions \cite{neurohive} called ResNets.

Sometimes high accuracy can be obtained by implementing CNN models with \textit{Residual Blocks}. Such models (ResNets) can be available in \textit{PyTorch}, \textit{TensorFlow} or \textit{Keras} \cite{neurohive} libraries \cite{residual}. ResNets are widely used for a variety of purposes. For example, in the context of the Covid-19 pandemic, there is a sudden surge in demand for fast and reliable diagnostic methods. A group of MIT researchers suggested an approach whereby artificial neural networks detect patients with Covid-19 based on the audio record of the cough \cite{mit_news}. The researches used a several neural networks for different subtasks, inter alia, they applied ResNet50 to discriminate between various sounds connected with a degree of vocal cord strength. The suggested model demonstrated remarkable results, as it succeeded to identified 98.5 \% of sick with Covid-19 patients even if they were completely asymptomatic. 

Another example of off-the-shelf residual models application is described in the article \cite{fingerprints}, where authors use ResNet50 to solve the fingerprint recognition problem. ResNet50 enabled the researchers to outrun the already existing solutions. Finally, with respect to the human activity recognition problem, ResNets can be applied as in the work \cite{transport}, where ResNet50 serves for the vision-based recognition of human activity.

Note that though off-the-shelf ResNet models usually tend to perform excellent results, there is no examples of their application to solve the acceleration-based human activity recognition problem as ResNets were developed for image classification while accelerometer signal is one-dimensional. 
%The reason is that off-the-shelf ResNets has a limit on minimal size of the input shape (namely, its width and height should not be smaller than 32 pixels). Consequently, ResNets are not able to deal with the raw accelerometer signal as it not even an image and so its width is equal to 1. 
We managed to overcome this obstacle by applying continuous wavelet transform to the initial signal and thus obtaining an image, as will be covered below.

Thus, though existing methods are getting pretty good with the HAR problem, they are still can be improved. Hence, we suggest a principally new approach for the challenge in question. Although CNNs rather can cope with 1D signal classification, while sorting 2D data they usually demonstrate noticably better results, so one-dimensional accelerometer signal, this signal has to be transformed. There are two the most frequently used ways to do it: \textit{Fourier transform} and \textit{wavelet transform} \cite{comparison}.
The Wavelet Transform method, suggested in 1970s by a French geophysicist Jean Morlet, enables to effectively obtain high resolution in frequency as well as in time domain. Since then the WT method has been significantly developed and in 2017 The Abel Prize was awarded to Yves Meyer “for his pivotal role in the development of the mathematical theory of wavelets” \cite{abel}.
Due to the periodicity of trigonometric functions and aperiodicity of mother wavelets, the Fourier transform is localized in frequency and is not localized in time and the Wavelet transform is localized in both of these domains. Consequently, the wavelet transform is much more suitable for analyzing non-stationary data (and the accelerometer signals belong precisely to the non-stationary one). 

%as while the Fourier transform is localized in frequency and is not localized in time, the Wavelet transform is localized in both of these domains. The periodicity of trigonometric functions and aperiodicity of mother wavelets is what account for this difference.

Therefore, in this paper we propose a model which firstly applies wavelet transform to signals and then classifies them via a convolutional neural network. We suggest that this approach will deliver quite higher accuracy than previously used methods.  In order to train and test our model we use UniMiB SHAR dataset. To evaluate the model we use four metrics: loss-function, accuracy, recall and precision.
The article is structured as followed. Firstly, we describe our program and the methodology of the experiment conduction. Then, the results of the experiment are given. Finally, we discuss the obtained results and the perspectives of the method. 

\section{Methodology}

This section describes the dataset we used to train and test the neuronal network, preprocessing and the network architecture.

\subsection{Data Description}

In this article, all the work is conducted with the UniMiB SHAR dataset. UniMiB SHAR is a publicly available dataset which was desired to estimate new methods of human activity analysis. UniMiB SHAR is a collection of accelerometer data obtained via smartphones that had been producted not earlier than in 2012. The process of accelerometer recording was as follows. Test subjects placed smartphones into their front trousers pocket and then performed the required actions. Before performing a movement and after finishing it, a test subject claps his hands in order to indicate start and end of the movement. From the accelerometer data the acceleration due to gravity was deducted. Finally, the Butterworth filter with the bandwidth up to 0,3 Hz was applied to the signal.

\begin{figure}[h]
    \centering
    \includegraphics[scale=1]{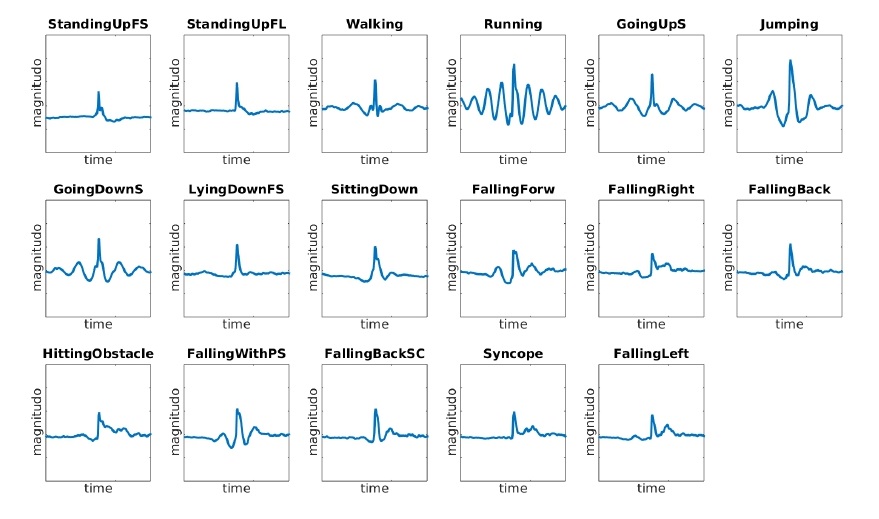}
    \caption{The averaged samples of acceleration shapes \cite{UniMiB}. Every subplot represents the magnitude of acceleration as a function of time}
    \label{pic1}
\end{figure}

Every datum in the dataset consists of the accelerometer values along three axes and the signal magnitude. The types of activities are firstly divided into two main parts, namely Activities of Daily Living (or ADLs) and Falls. ADLs includes five categories of activities: Context-related (e.g., Stepping in a car), Motion-related (e.g., Walking), Posture-related (e.g., Standing), Sport-related (e.g., Jumping)  and Others (e.g., Vacuuming). Falls are also divided into categories: Falling backward, Falling forward, Falling sideward and Specific fall (Fig. \ref{pic1}). Therefore, there are 4 subsets: AF2, F8, A9 and AF17. AF2 is divided only on Falls and ADLs. F8  contains 8 classes of Falls, A9 contains 9 classes of ADLs. Finally, AF17 is composed of all the 17 classes. 

In the current article we work with the A17 subset only. We shuffle the dataset and then randomly leave 0,8 of data for training and the rest 0,2 for test. 

%We randomly left 0,2 of data for test, having the dataset previously shuffled.

\newpage
\newpage

\subsection{Fourier and Wavelet Transforms}

\subsubsection{Feature extraction}

Signal analysis is a popular field of research at the interface between mathematics and radio engineering. In order to present data in an interpretive way, one or the other approach to feature selection are usually implemented. For example, in the work \cite{records} the authors tried to apply 35 parameters such as signal energy $E$ or the entropy $EE$ of the energy of each frame $E_n$ to a signal $S$:

\begin{equation}
    E = 
    \sum_{t=1}^T
    |S_t|^2,
    \label{eq_energy}
\end{equation}

\begin{equation}
    EE = 
    - \sum_{n} \frac{E_n}{E} ln \frac{E_n}{E}.
    \label{eq_energy}
\end{equation}

Such approaches can leads to agreeable results, though sometimes the feature extraction by the signal decomposition is required. For this purpose there is a whole series of methods. We cover them in the further subsections. 

\subsubsection{Generalized Fourier Series}

Generalized Fourier series is the method which is used the most commonly for the signal analysis problem. 
A signal $S(t)$ which satisfies the equation

\begin{equation}
    \int_{t_1}^{t_2}[S(t)]^2dt < \infty
    \label{eq0}
\end{equation}

\noindent can be decomposed by the basis functions $\phi_0(t),...,\phi_n(t)$ via the \textit{Generalized Fourier series}:

\begin{equation}
    S(t) = \sum_{n=0}^{\infty}C_n\phi_n(t),
    \label{eq01}
\end{equation}

\noindent where 

\begin{equation}
    C_n = \frac{1}{||\phi_n||^2}\int_{t_1}^{t_2}S(t)\phi_n(t)dt,  \; \;
    ||\phi_n||^2 = \int_{t_1}^{t_2}\phi_n^2(t)dt.
    \label{eq001}
\end{equation}

\if 0
Expression (\ref{eq01}) is called \textit{Generalized Fourier series}. The coefficients of decomposition $C_0,...,C_n$ are calculated as 

\begin{equation}
    C_n = \frac{1}{||\phi_n||^2}\int_{t_1}^{t_2}S(t)\phi_n(t)dt,
    \label{eq02}
\end{equation}

\noindent where $||\phi_n||^2$ is the norm of the function $\phi_n(t)$ and is defined as
\begin{equation}
    ||\phi_n||^2 = \int_{t_1}^{t_2}\phi_n^2(t)dt.
    \label{eq03}
\end{equation}

\fi

Basis functions $\phi_0,...,\phi_n$ should not be identically zero and, besides, they should be orthogonal on the interval $[t_1, t_2]$:

\begin{equation}
    \forall \; k, \;n \; (k \neq n): \; \;
    \int_{t_1}^{t_2}\phi_k(t)\phi_n(t)dt = 0
    \label{eq04}
\end{equation}

If every basis function $||\phi_n||^2 = 1$ (such basis functions are called normalized), the basis is called orthonormal. The choice of the optimal orthonormal system of basis functions is a matter of great importance. Two the most frequently used systems are harmonic basis functions and wavelets. 

\subsubsection{Fourier Transform}

\textit{Fourier transform} allows to decompose a signal by harmonic functions $1$, $sin$ $x$, $cos$ $x$, $sin$ $2x$, $cos$ $2x$,..., $sin$ $nx$, $cos$ $nx$ \cite{kr}. 

A periodic signal can be decomposed into \textit{Fourier Series} as follows:

\begin{equation}
    S(t) = \frac{a_0}{2} + \sum_{n=1}^{\infty}(a_n cos (nt) + b_n sin(nt)),
    \label{eq05}
\end{equation}

\noindent where 

\begin{equation}
    a_0 = \frac{1}{\pi}\int_{-\pi}^{\pi}S(t)dt, \; \;
    a_n = \frac{1}{\pi}\int_{-\pi}^{\pi}S(t)cos(nt)dt, \; \;
    b_n = \frac{1}{\pi}\int_{-\pi}^{\pi}S(t)sin(nt)dt.
    \label{eq005}
\end{equation}

If the signal S(t) is aperiodic, it can be represented as \textit{Fourier Integral} via applying \textit{Fourier Transform}:

\begin{equation}
    S(t) = \int_{-\infty}^{\infty}[A(\omega)cos(\omega t) + B(\omega)sin(\omega t)]d\omega, 
    \label{eq05_0}
\end{equation}

\noindent where $A(\omega)$ and $B(\omega)$ are defined likewise $a_n$ and $b_n$:

\begin{equation}
    A(\omega) = \frac{1}{\pi}\int_{-\infty}^{\infty}S(t)cos(wt)dt, \; \;
    B(\omega) = \frac{1}{\pi}\int_{-\infty}^{\infty}S(t)sin(wt)dt.
    \label{eq05_00}
\end{equation}

\begin{figure}[h]
    \centering
    \includegraphics[scale=0.5]{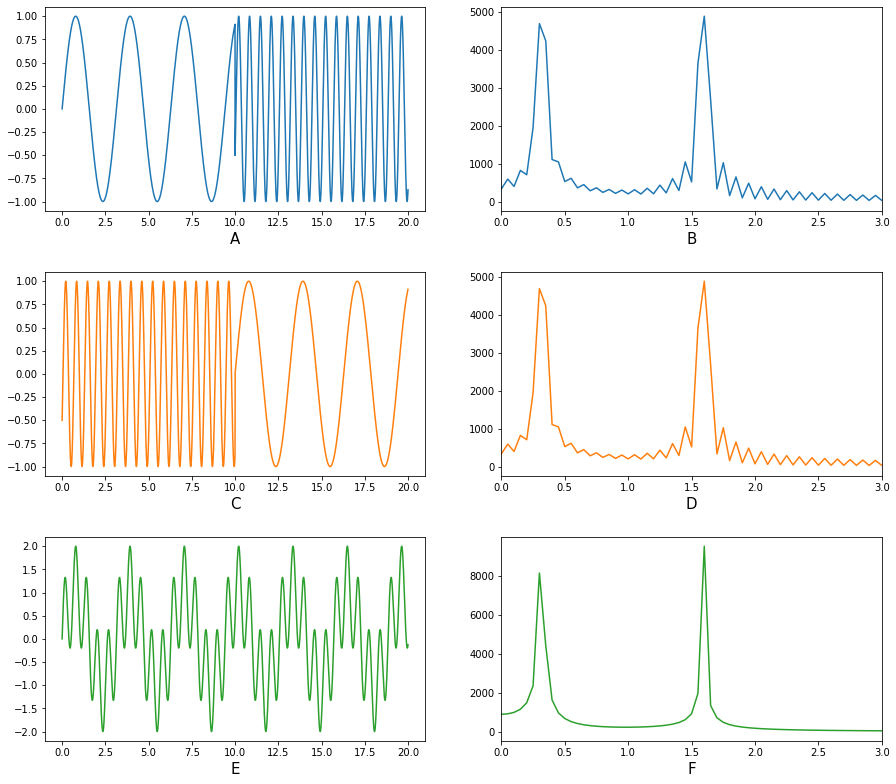}
    \caption{Different functions and Fourier transform applied to them. \textit{A:} $y = sin \: 2x, x<10$ and $y = sin \: 10x, x>10$. \textit{B:} $y = sin \: 10x, x<10$ and $y = sin \: 2x, x>10$. \textit{C:} $y = sin \: 2x + sin \: 10x$). \textit{D}, \textit{E}, \textit{F:} Fourier transforms applied to functions \textit{A}, \textit{B} and \textit{C} respectively. Thus, distinctly different functions can have very similar Fourier transforms}
    \label{pic1_0}
\end{figure}

Fourier transform enables to reveal the intensity of each frequency which the original signal consists of, however, the time localizations of these frequencies remain unknown \cite{from}. Due to the unability to localize frequencies in the time domain, Fourier transform is unsuitable for analyzing non-stationary signals. For example, Fig. \ref{pic1_0} demonstrates how substantially different signals can have remarkably similar Fourier   transforms. There are three signals in the first column (\textit{upper}: $y = sin \: 2x, x<10$ and $y = sin \: 10x, x>10$; \textit{center}: $y = sin \: 10x, x<10$ and $y = sin \: 2x, x>10$; \textit{lower}: $y = sin \: 2x + sin \: 10x$). The second column consists of the Fourier transforms applied to the signals from the first column. The Fourier transform is obtained via using the \textit{scipy.fft} module.

\subsubsection{Short-Time Fourier Transform}

One of the ways to specify the frequency location is to decompose the original signal only in a certain window which glides alone the time axis. This approach is known as \textit{Short-Time Fourier Transform} \cite{short}. It can be represented as follows:

\begin{equation}
    STFT(t, \omega) = \int_{-\infty}^{\infty}S(\tau)W(\tau - t)e^{-iw\tau},
    \label{eq06}
\end{equation}

\noindent where $W(\tau-t)$ is the window function. The type of the window function and the window size are chosen for every certain purpose individually. It should be noted, that the size of the window imposes some restrictions on the quality of results. According to the uncertainty principle \cite{uncertainty}, with higher frequency resolution the time resolution decreases and vice versa:

\begin{equation}
    \Delta \tau \cdot \Delta \omega \sim %\geqslant
    \frac{1}{4\pi}.
    \label{eq07}
\end{equation}

Sometimes it is possible to find a compromise window size (for example, in the article \cite{har_short} Short-Time Fourier Transform is successfully used to solve the HAR problem), but in certain cases it becomes necessary to find some other approach free of this restriction. This disadvantage can be wiped out by implementing the \textit{Wavelet Transform} approach.

\subsubsection{Wavelet Transform}

There are two types of wavelet transform: \textit{discrete} and \textit{continuous} \cite{polikar}. 

\textit{Continuous Wavelet Transform} (or \textit{CWT}) is a mathematical operation which represents a real-valued function $S(t)$ as a following integral:

\begin{equation}
    W_s(a, b) = \frac{1}{\sqrt{a}}\int_{-\infty}^{\infty}S(t)\Psi \left(\frac{t-b}{a}\right)dt,
    \label{eq1}
\end{equation}

\noindent depending on a scale $a > 0$ ($a \in \mathbb{R}^+$) and translocational value $b$ ($b \in \mathbb{R}$). Discrete Wavelet Transform (or DWT) carries with it a similar idea  \cite{marry}, with the difference that parameters $a$ and $b$ are discrete:

\begin{equation}
    a = (a_0)^n, \; \;
    b = kb_0.
    \label{eq1_00}
\end{equation}

Though CWT and DWT have much in common, they are usually used for different purposes. While DWT is a perfect instrument for such coding problems as image compression, CWT is mostly applied for signal analysis tasks \cite{fast}. Thus, CWT is the method we implement in the current work.

\begin{figure}[h]
    \centering
    \includegraphics[scale=0.5]{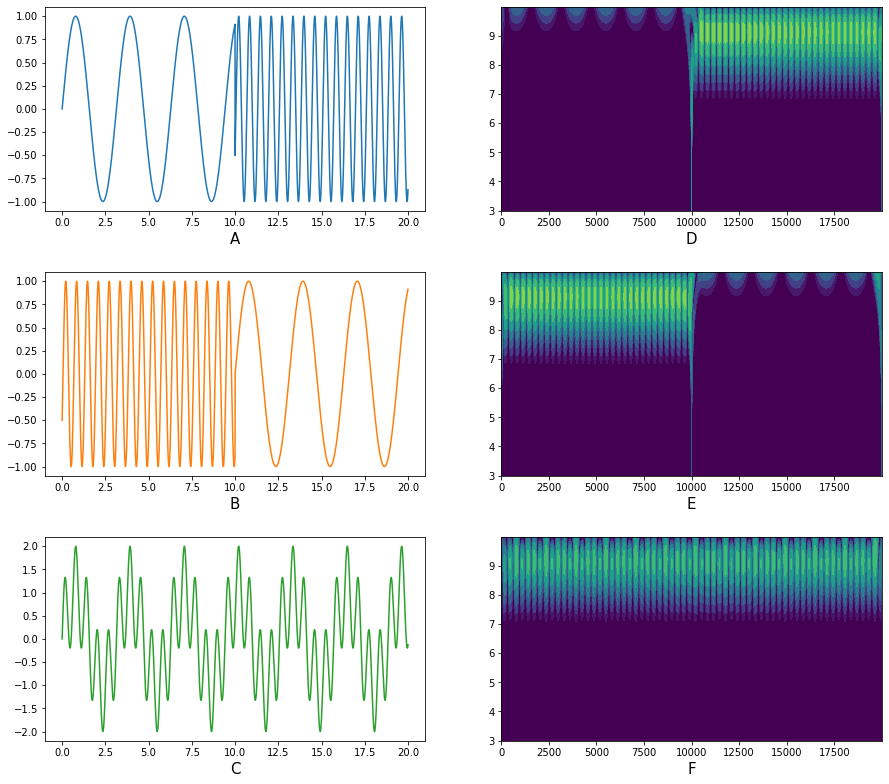}
    \caption{Continuous wavelet transform with the Mexican Hat mother wavelet applied to different functions. \textit{A}, \textit{B}, \textit{C}: the functions identical to those in Fig. \ref{pic1_0}. \textit{D}, \textit{E}, \textit{F}: wavelet transforms of functions \textit{A}, \textit{B} and \textit{C} respectively}
    \label{pic1_0_1}
\end{figure}

The Fig. \ref{pic1_0_1} demonstrates that compared to the Fourier transform (Fig. \ref{pic1_0}), the Wavelet transform is able to distinctly localize features in the time domain.

As it has already been mentioned, CWT provides an excellent opportunity to extract and investigate complicated spectral features of a signal \cite{sad}. Function $\Psi$ is a continuous in time and frequency function and is called mother wavelet. This mother wavelet is used to obtain a daughter wavelet for each possible pair ($a$, $b$):

\begin{equation}
    \Psi_{a, b}(t) = \frac{1}{\sqrt{a}}\Psi\left(\frac{t-b}{a}\right).
    \label{eq1_0}
\end{equation}

Then, the CWT is applied:

\begin{equation}
    W_s(a, b) = \frac{1}{\sqrt{a}}\int_{-\infty}^{\infty}S(t)\Psi
    \left(\frac{t-b}{a}\right)dt 
    = \int_{-\infty}^{\infty}S(t)\Psi_{a, b}(t) = (S(t), \Psi_{a, b}(t)).
    \label{eq1_1}
\end{equation}

Formula (\ref{eq1_1}) shows the similarity between the signal in question and each of the daughter wavelets. These results can be represented as an image with $b$-value set along the $x$-axis and $a$-value set along the y-axis. The intensity of each pixel is determined by formula \ref{eq1} with corresponding $a$ and $b$ values.

\begin{figure}[h!]
    \centering
    \includegraphics[scale=0.57]{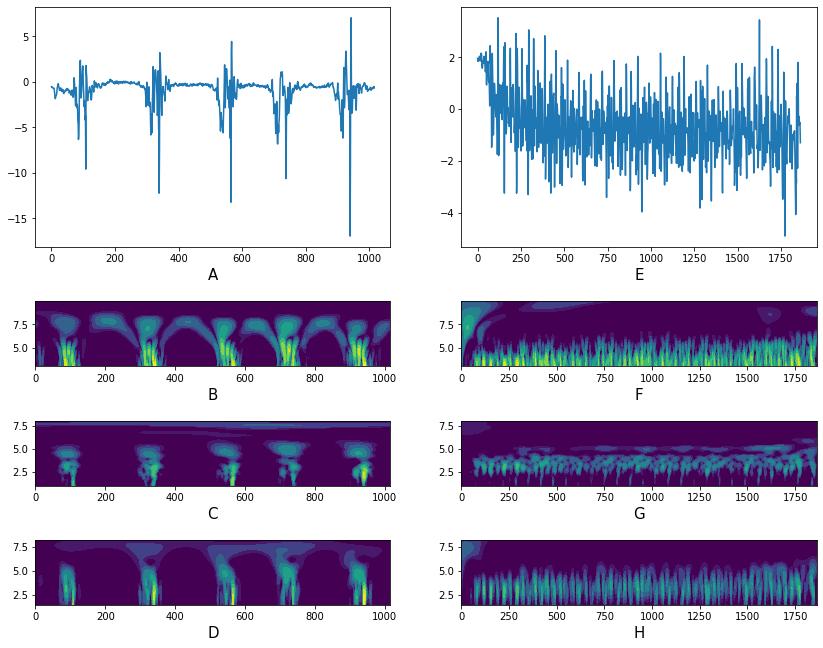}
    \caption{The result of continuous wavelet transform (CWT) for Jumping and Running signals. \textit{A}: acceleration along the X-axis for \textit{jumping}. \textit{B}, \textit{C}, \textit{D}: wavelet transform for signal \textit{A} for \textit{Mexican Hat}, \textit{Morlet} and \textit{Paul} mother wavelets respectively. \textit{E - H}: analogous signal and wavelet transforms for \textit{walking}}
    \label{pic2}
\end{figure}

In the fig. \ref{pic2} the results of CWT applied to the \textit{jumping} and \textit{walking} signals using different mother wavelets.

It has already been said that mother wavelet $\Psi$ has to be continuous in time and frequency. There are 3 more requirements for mother wavelet functions. First, such function must be \textit{limited} what means that its squared module has to be limited:

\begin{equation}
    || \Psi ||^2= 
    \int_{-\infty}^{\infty}||\Psi (t)||^2dt < \infty
    \label{eq2}.
\end{equation}

\noindent Second, the function has to be \textit{localized} both in time and in frequency. Finally, the area under the curve has to be \textit{zero} \cite{yakovlev}. Let us address the last requirement in greater detail. In other words, the zero-order moment of the function should be equal to zero \cite{zero}:

\begin{equation}
    \int_{-\infty}^{\infty}\Psi(t)dt = 0.
    \label{eq2_0}.
\end{equation}

There are several widely-used mother wavelets, but we chose three of them: \textit{Morlet}, \textit{Paul} and \textit{derivatives of gaussian function} (including \textit{Mexican Hat}, which is second-order derivative). In the current work the wavelet transform is performed via the \textit{pycwt} python library.

\begin{figure}[h!]
    \centering
    \includegraphics[scale=0.55]{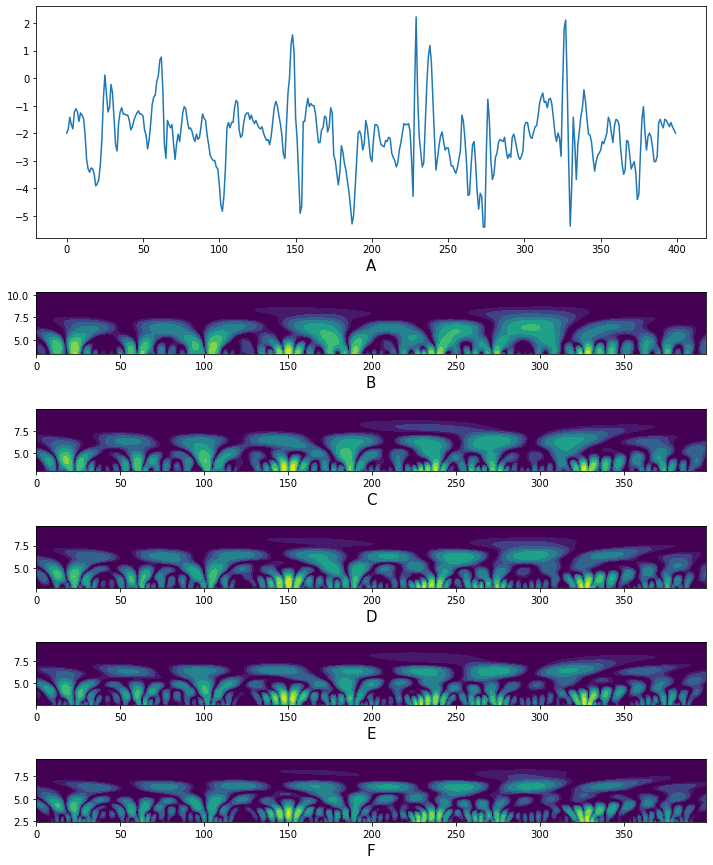}
    \caption{CWT for different number of zero moments. \textit{A}: Acceleration along the X-axis for \textit{GoingDownS}. \textit{B - F}: wavelet transforms for signal \textit{A} with \textit{DOG} mother wavelets with \textit{1}-\textit{5} zero moments respectively}
    \label{pic2_0}
\end{figure}

Sometimes there are additional conditions on the function $\Psi(t)$:

$$\forall k \in [0, n]:$$
\begin{equation}
    \int_{-\infty}^{\infty}t^k\Psi(t)dt = 0,
    \label{eq2_1}.
\end{equation}

\noindent which mean that not only zero-ordered moment but all the moments up to one with the n-order are equal to zero. Wavelets with low orders of zero moments are targeted at low frequencies primarily (pic. \ref{pic2_0}). On the other hand, high-ordered wavelets mitigate lowly oscillating components of signal and thus reveal high-frequency structures. Hence, a combination of low and high-ordered filters can present the most complete picture of signal features.

Applying CWT is a popular approach in the different branches of the signal classification field. For example, in the work \cite{automatic} CWT is used for the microseismic signals classification and in the article \cite{cardio} it is an excellent tool for dealing with electrocardiogram signals.

In the current work we apply CWT not only because it is an excellent tool for feature extraction, but also to transform a single-dimensioned signal into an image. This step is necessary, since convolutional neural networks, which we use, tend to provide significantly better results on 2- or 3-dimensional objects, as will be covered in the next section. Therefore, convolutional wavelet transform opens up new horizons of signal classification, inter alia, in the sphere of human activity recognition.

\newpage

\subsection{Convolutional Neural Network}

Convolutional Neural Networks (or CNNs) is a commonly used class of feedforward learning algorithms \cite{dcnn}. The main idea of CNN working is a shift from minor features to the global ones, what makes CNNs perfect for image classification. A typical CNN consists of alternating batches of convolutional, pooling and fully connected layers. Convolutional layers are arranged as follows. A small matrix called filter courses through a previous layer and after each shift the filter is multiplied componentwise by a part of that layer; the next layer is being formed from the results of these multiplications. Convolutional layers are designed for feature extraction, and every further convolutional layer derives more complex features than the previous one. Pooling layers are responsible for the reduction of feature size. A pooling layer takes a square of a few pixels and returns the average (average pooling) value or the maximal value (max pooling) of them. The last type of layers applied in CNNs is fully connected layers. Their main purpose is to present a function of class determination based on features extracted by all the previous layers.

\newpage

\begin{figure}[h]
    \centering
    \includegraphics[scale=0.45]{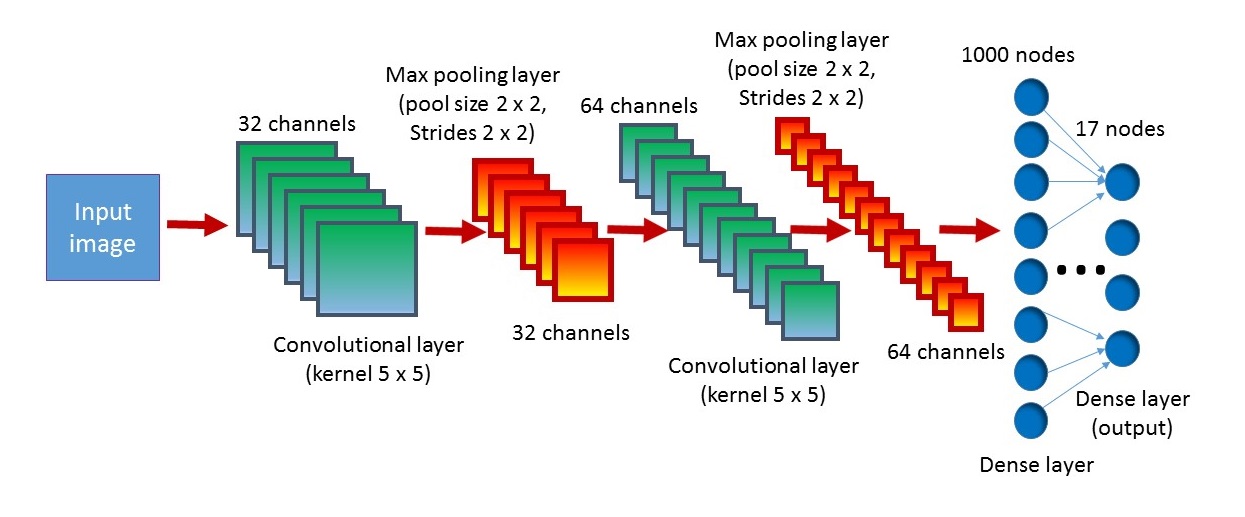}
    \caption{The principal scheme of the model}
    \label{pic3}
\end{figure}

The initial version of our CNN consists of alternating two convolutional (kernel size = $5*5$ and two  max pooling layers (2-fold pooling in both directions), followed by two dense layers (first of them has 1000 neurons and the second one has 17 neurons, 
what is equal to the number of activity classes). We have the ReLu activation function in the convolutional layers and in the 1000-neuroned dense layer and the Softmax activation function in the last dense layer.The stride (a kernel step) is equal to 1 in the convolutional layers and to 2 in the pooling layers. The principal scheme of the initial model is shown in the fig. \ref{pic3}. Some parameters (such as the number of convolutional layers, fully-connected (dense) layers or the number of neurons in a layer) are varied in the course of the work.

Deep CNNs are able to extract more implicit features, however, in deep networks the vanishing gradient problem may occur. This matter can be solved by using \textit{Residual Blocks} \cite{He_k}, which add shortcut connections between pairs of layers. The function of residual block is as follows:

\begin{equation}
    y = \mathcal{F}(x, W_i)+x,
    \label{resblock}
\end{equation}

\noindent where $x$ and $y$ are the input and output vectors respectively and function  $\mathcal{F}(x, W_i)$ refers to the residual mapping. In the current article we implement ResNets available in the \textit{Python} library \textit{Keras}.

The general workflow of our work is as follows (Fig. \ref{pic1_TOC}). Though for the 1D signal classification CNNs are also suitable, like they are implemented in the article \cite{earth} for the seismic signal classification, while dealing with 2D objects CNNs can perform significantly better results. Thus, firstly, we convert the 1D accelerometer signal into the 2D images via applying CWT in order to extract signal features and, at the same time, to make it possible to implement 2D CNNs. Secondly, these images are cropped down to the similar size and converted to the grayscale for reducing excessive dimension. The mentioned steps are conducted for three spacial axes ($x$, $y$ and $z$) and then the obtained gray images are combined into a multi-axis image. Finally, the multi-axis images are received by a neural network (CNN or ResNet), which predicts the class of the activity.

%\newpage

\newpage
\section{Results and Discussion}

\subsection{Preprocessing}

Firstly, we choose an optimal way of data representation. We compare the image methods, where the image representation is realised via three different python libraries: \textit{matplotlib.pyplot}, \textit{cv2} and \textit{skimage}, and the array method. We build a primitive convolutional neural network (Fig. \ref{pic3}), which gives the best results while using the images created via the \textit{matplotlib.pyplot} library. Thus, we chose this way to create images as the most appropriate way of data representation.

%\subsubsection{libraries for image saving}

%Надо ли оставить этот подпункт?

%The first step taken in the following research was to choose an optimal way of data representation. We compared the image methods, where the image representation were realised via three different python libraries: \textit{matplotlib.pyplot}, \textit{cv2} and \textit{skimage}, and the array method. All the images were converted to a grayscale. The pixels of \textit{cv2} images were also divided by 256 for the purpose of normalization. To estimate the suitability of the four mentioned ways, we built a primitive convolutional neural network consisted only of alternating two convolutional and two maxpooling layers, which were followed by a pair of dense layers (сюда еще можно вставить небольшой workflow для описанной сетки).  The resulted loss functions and accuracies are presented in the fig. \ref{pic6}. The best results were obtained while using the images created via the \textit{matplotlib.pyplot} library. Thus, we chose this way to create images as the most appropriate for our purpose way of data representation.

\newpage
\subsection{Different axes}

\begin{figure}[h]
    \centering
    \includegraphics[scale=0.6]{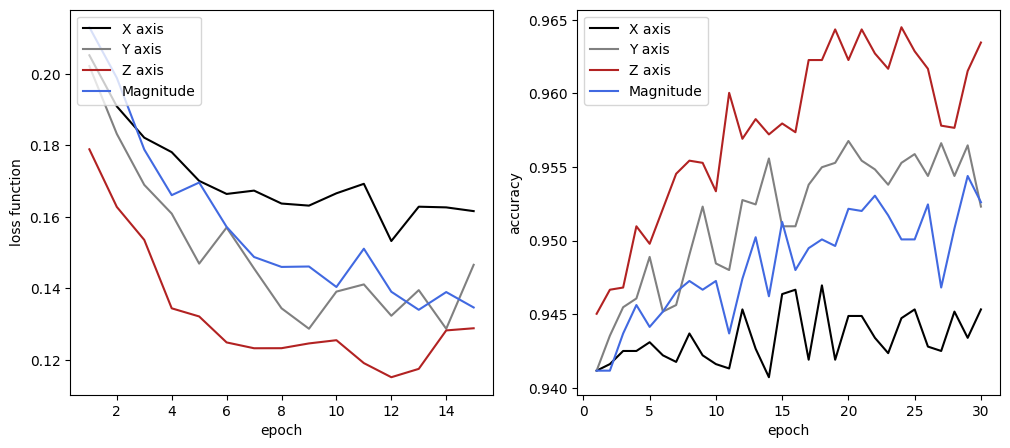}
    \caption{Loss-function and accuracy vs number of epochs for magnitude and different axes}
    \label{pic7}
\end{figure}

As stated above, in this paper we work with the \textit{UniMiB SHAR dataset}. It contains acceleration values along three axis and general magnitude of acceleration. Using the same network (Fig. \ref{pic3}), we investigate which one of these four data series characterizes activity most distinctly. Then we test if a combination of data rows can deliver better results. The results are presented in Fig. \ref{pic7} (here and throughout all the plots are for the test data). It can be seen that Z axis characterizes the general motion most closely.

%Next, we investigated wavelet images obtained with acceleration projected on different axes and with general magnitude of the acceleration.

\begin{figure}[h]
    \centering
    \includegraphics[scale=0.6]{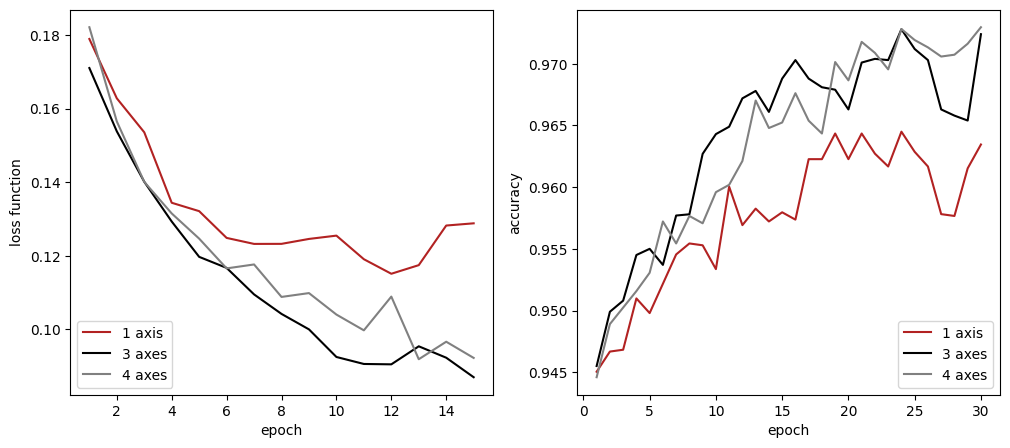}
    \caption{Loss-function and accuracy vs number of epochs for 1, 3 and 4-axis neural networks}
    \label{pic8}
\end{figure}

After that we designed 3-axis (X, Y, Z) and 4-axis (X, Y, Z, magnitude) neural networks in order to find out if the combination of axes describes motions most specifically. It can be seen from Fig. \ref{pic8} that in comparison with single-axis neural network (Z axis), multi-axis neural networks produce substantially better results. On the other hand, the results obtained via 3-axis and 4-axis neural networks do not differ from one another significantly. Meanwhile,  3-axis neural network is less time- and resource-consuming than the 4-axis one, so we chose 3-axis network.

%and if the result can be improved even more by adding the magnitude.The loss-functions and accuracy functions obtained on multi-channelled neural networks were compared with loss-function and accuracy obtained on single-channelled neural network trained on Z axis acceleration wavelets (it is Z axis which was chosen as its metrics turned out to be the best according to the previous experiment). It can be seen from fig. \ref{pic8} that in comparison with single-channelled neural network, multi-channelled neural networks produce substantially better results. On the other hand, the results obtained via 3-channelled and 4-channelled neural networks do not differ from one another significantly, so we decided to choose 3-channelled neural network as it is less time-consuming than the 4-channelled one. (тут еще можно сказать про то, что модель не переобучаема, и вставить соответствующий график [для обучающей и тестовой выборки на одном графике]).

\newpage
\subsection{Network variations}

Next, we optimize the network architecture for 3-axis wavelet images

%To obtain a more detailed characteristic of predictive quality, we decided to use not only accuracy and loss-function? but also recall and precision metrics.

\begin{figure}[h!]
    \centering
    \includegraphics[scale=0.6]{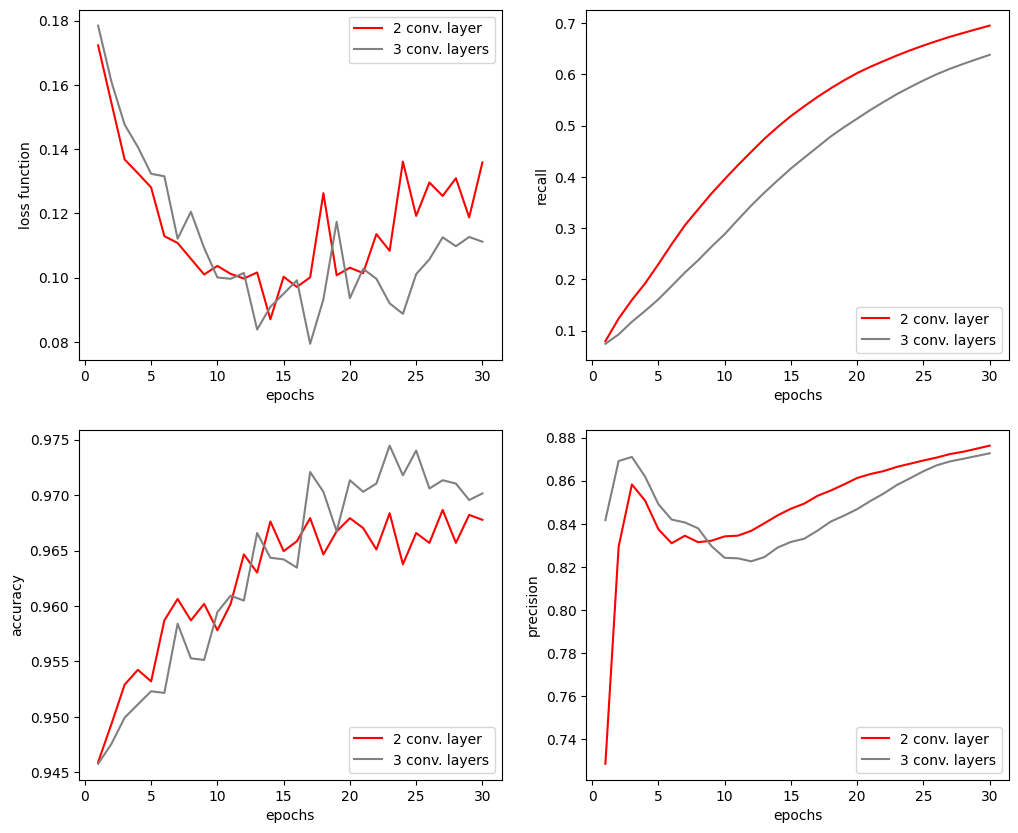}
    \caption{Loss-function, recall, accuracy and precision vs number of epochs for 2 or 3 convolutional layers}
    \label{pic9}
\end{figure}

To start with, we probe different numbers of convolutional layers. 
We find out, that if the net consists of 3 convolutional layers its predictive quality increases compared with a net consisted of 2 convolutional layers (Fig. \ref{pic9}).

\newpage

\begin{figure}[h!]
    \centering
    \includegraphics[scale=0.6]{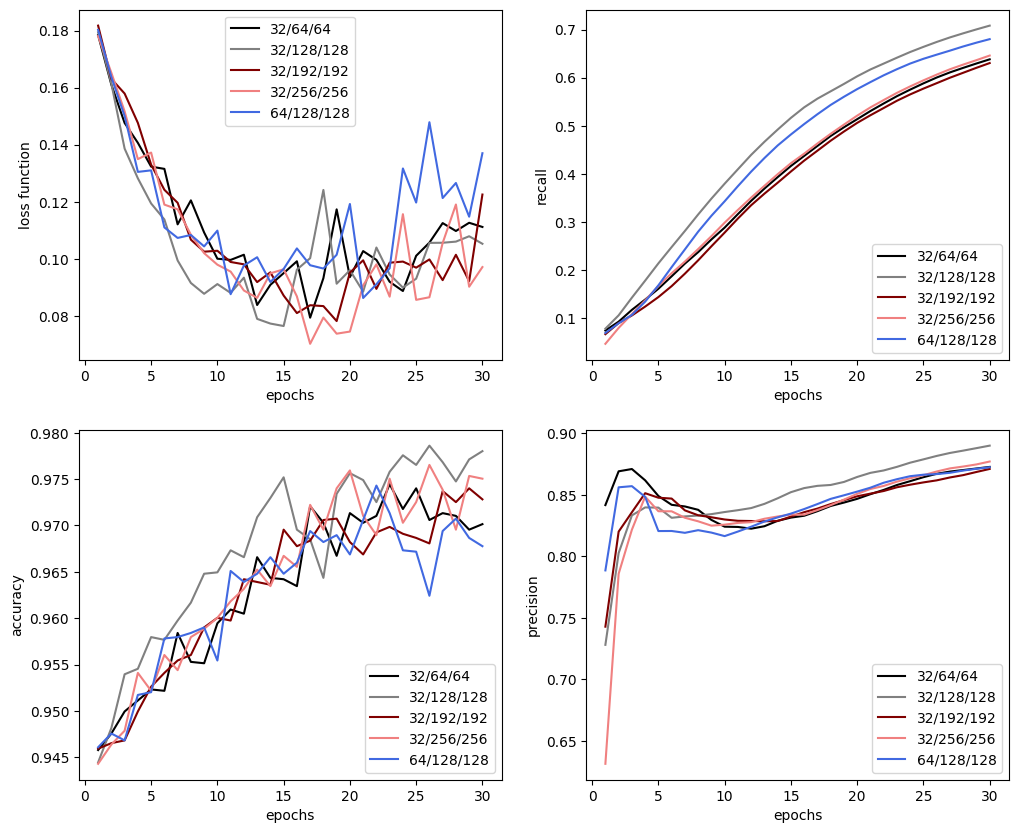}
    \caption{Loss-function, recall, accuracy and precision vs number of epochs for different number of neurons in convolutional layers}
    \label{pic10}
\end{figure}

Secondly, we examine what is the optimal number of neurons in each convolutional layer. It turns out to be 32/128/128 (fig. \ref{pic10}).

\newpage

\begin{figure}[h!]
     \centering
    \includegraphics[scale=0.6]{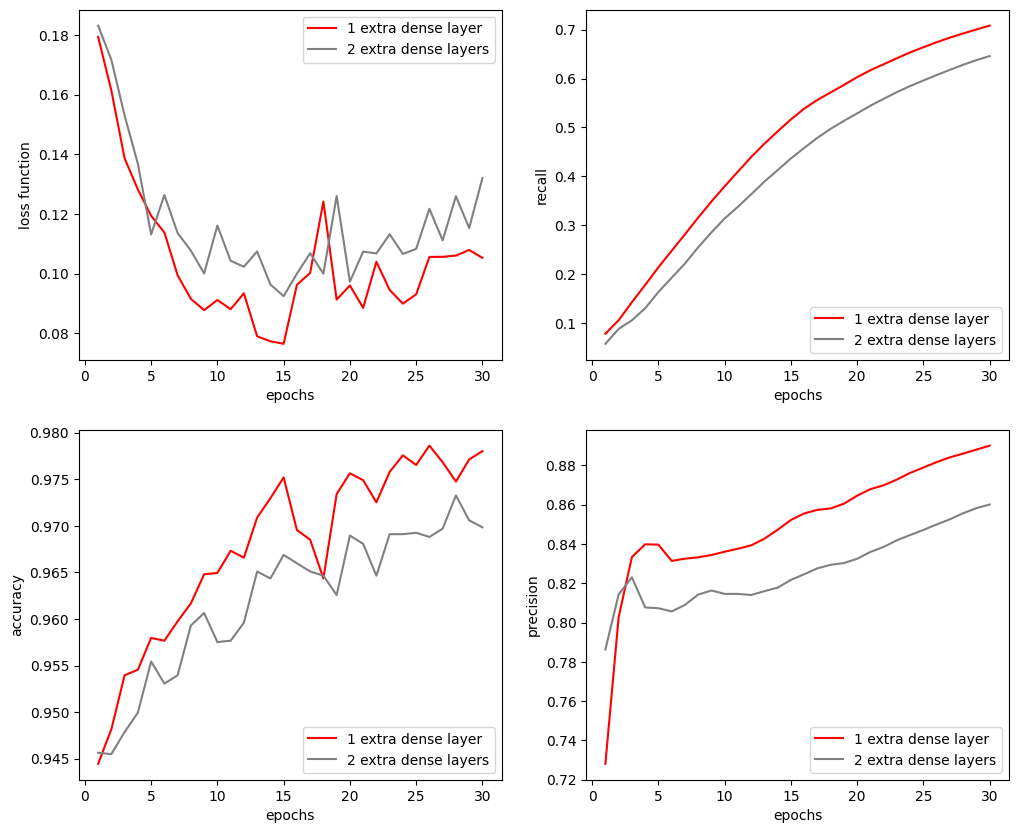}
    \caption{Loss-function, recall, accuracy and precision vs number of epochs for different number of extra dense layers}
    \label{pic11}
\end{figure}

Finally, for a net with 3 convolutional layers consisted of 32/128/128 neurons we determine the optimal number of extra dense layers. It is established that while the number of extra dense layers increases the accuracy only declines, as well as other metrics do. (Fig. \ref{pic11}).

%\newpage

%Finally, we tested if the adding of dense layers with a less number of neurons can improve the results, but this hypothesis failed (fig. \ref{pic12}). 

%\newpage

Thus, the most appropriate network is composed of 3 convolutional layers with 32/128/128 neurons and one extra dense layer with 1000 neurons. 

\newpage

\subsection{Cut and uncut images}

\begin{figure}[h!]
    \centering
    \includegraphics[scale=0.6]{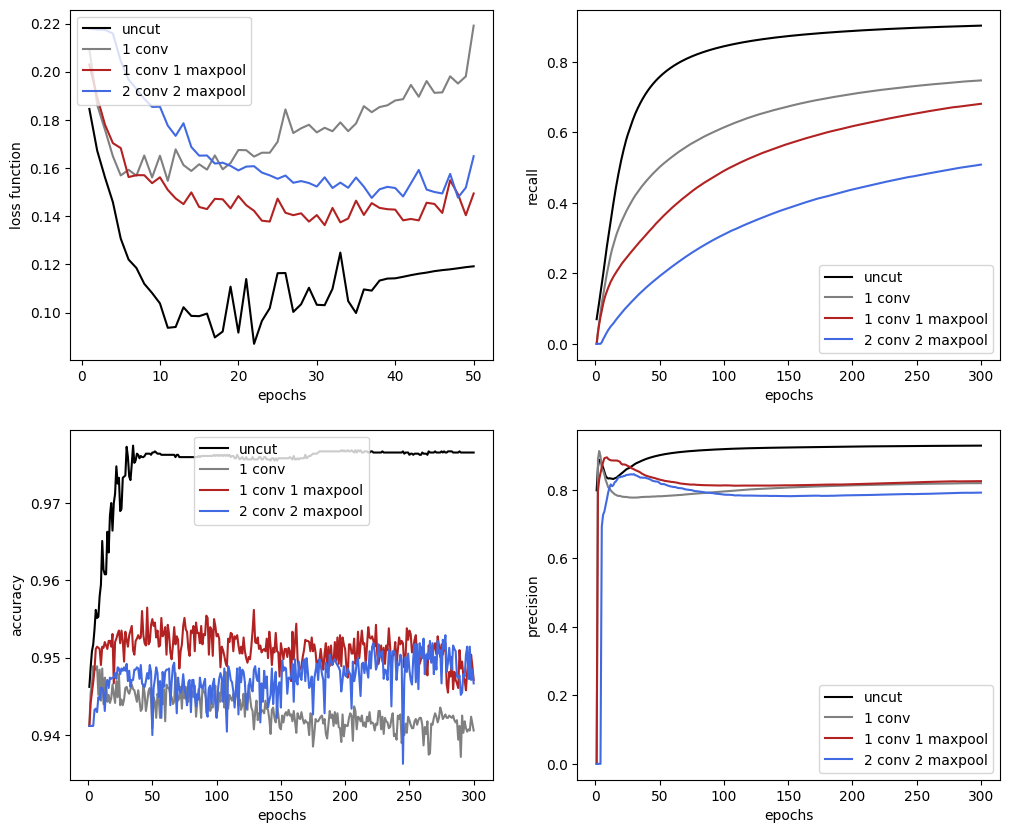}
    \caption{Loss-function, recall, accuracy and precision vs number of epochs for cut and uncut images}
    \label{pic13a}
\end{figure}

So far, we overlook the fact that wavelets are not just simple images with all the pixels equitable. Indeed, the informativity of values changes along the y axis, and in the area of low scale parameter even the least visible features can be detected. Thus, we try to divide every image into two parts as a branched network should be more sensitive, but as the number of trainable parameters decreases the result degradates (Fig. \ref{pic13a}).

\newpage
\subsection{Different wavelets}

\begin{figure}[h!]
    \centering
    \includegraphics[scale=0.6]{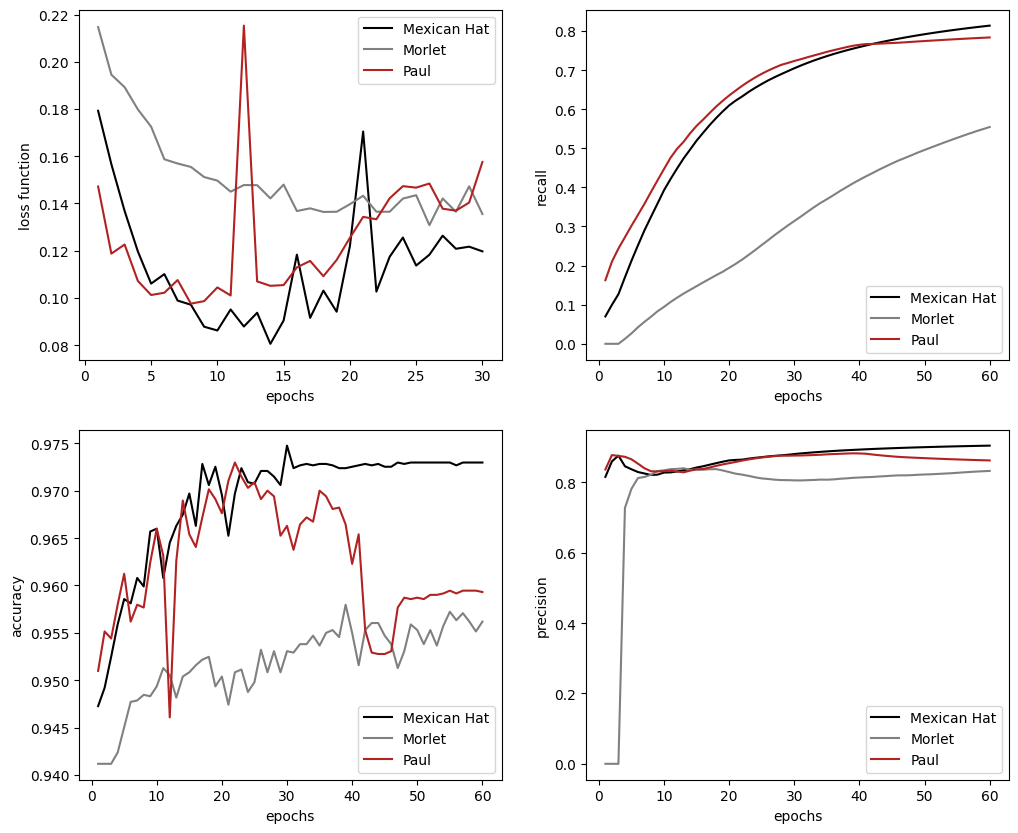}
    \caption{Loss-function, recall, accuracy and precision vs number of epochs for different mother wavelets}
    \label{pic14}
\end{figure}

The choice of the optimal mother wavelet is a very important step of the research. Up to this point, we use only the \textit{Mexican Hat} wavelet (the 2nd derivative of the \textit{Gauss function}). Now we try two more functions: \textit{Paul} and \textit{Morlet}. However, Mexican Hat shows the best scores (Fig. \ref{pic14}). A little less accuracy can be delivered by the Paul wavelet, and the Morlet Wavelet gives the worst result.

%\newpage

\newpage
Next, we explore combinations of several mother wavelets. We add new wavelets in order of accuracy decreasing: Mexican Hat, Paul and Morlet. 
%Asn the current work wthe pictures cae n be obtained via different mother wavelets, we evaluated investigated which mother wavelet deliver the best result.
%It turned out to be Mexican Hat wavelet (fig. \ref{pic13}). A little less accuracy can be delivered by the Paul wavelet, which is followed by the Morlet Wavelet and the Gaussian with the 0 degree of differentiation. This is the order which we added wavelets in general combination in. 

%\newpage

\begin{figure}[h!]
    \centering
    \includegraphics[scale=0.6]{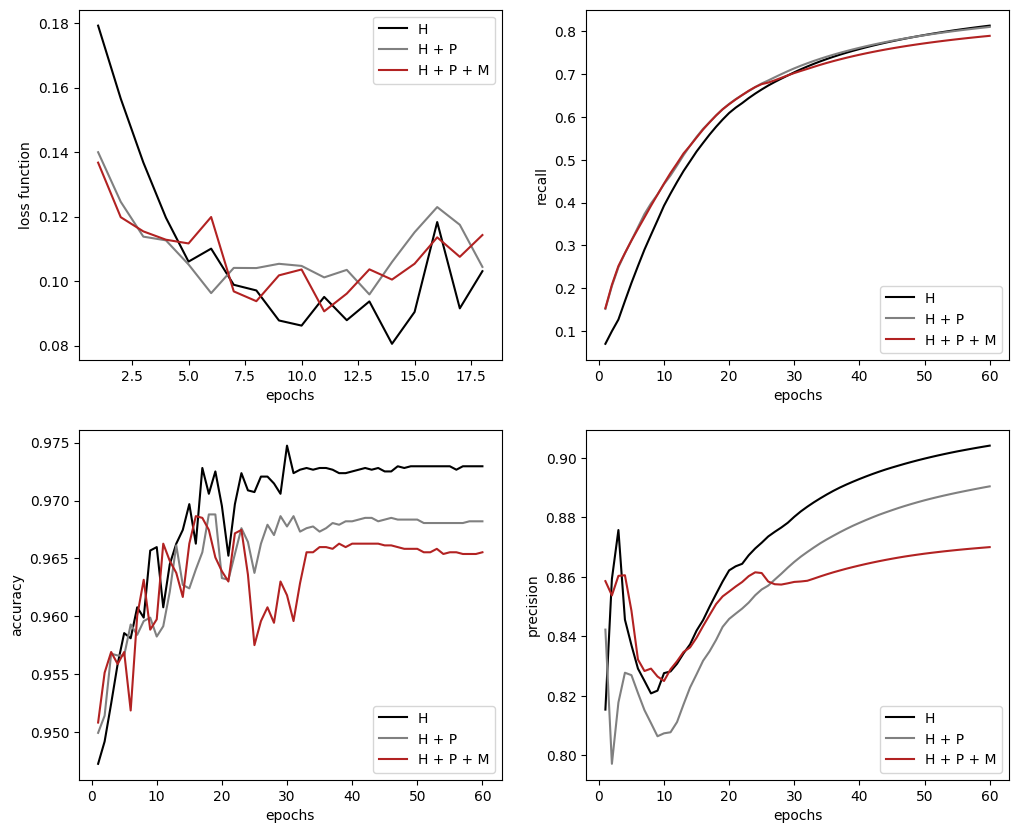}
    \caption{Loss-function, recall, accuracy and precision vs number of epochs for different cominations of mother wavelets (H, P and M are the Mexican Hat, Paul and Morlet mother wavelets respectively)}
    \label{pic15}
\end{figure}

It can be seen, that while the number of wavelets increases the accuracy decreases (Fig. \ref{pic15}), so we continue working with Mexican Hat only.

\newpage

\subsection{Batchsize}

\begin{figure}[h!]
    \centering
    \includegraphics[scale=0.6]{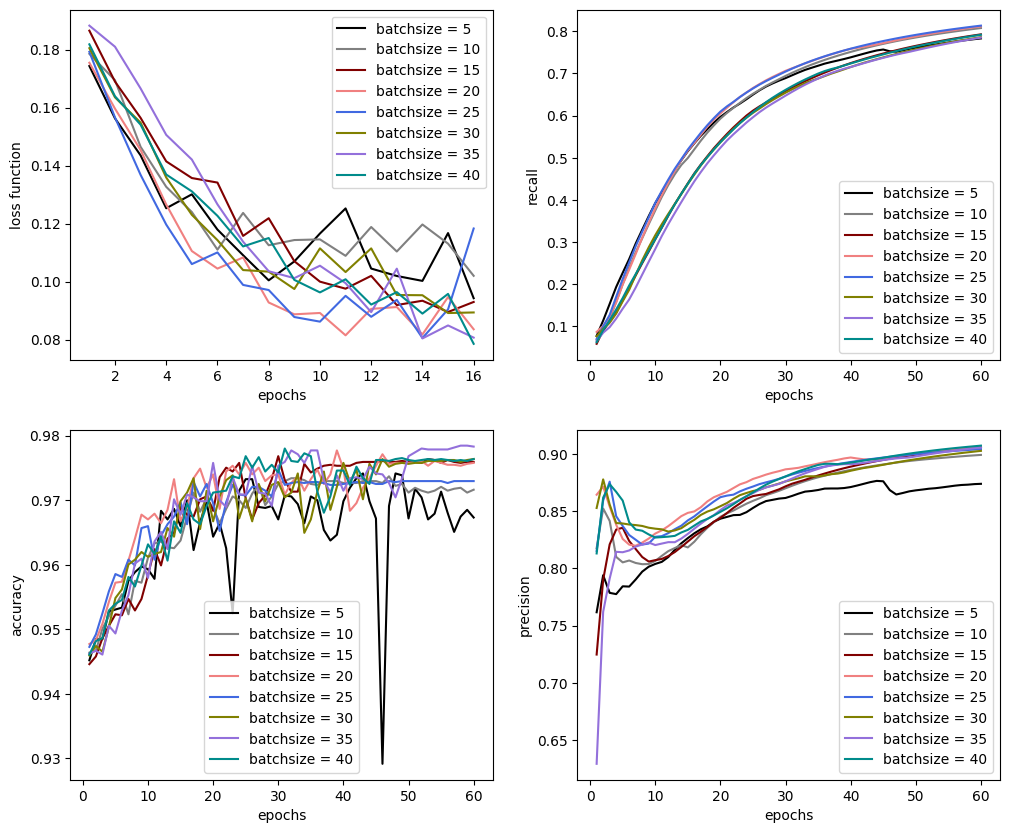}
    \caption{Loss-function, recall, accuracy and precision vs number of epochs for different number of batches for the Mexican Hat mother wavelet}
    \label{pic16}
\end{figure}

Now, we investigate what batchsize is the best for the chosen model (Fig. \ref{pic16}). For batchsize = 35 the best result is obtained (accuracy is equal to 0.978).

%Тут был еще раздел про обучение не на всех данных, но он какой-то бесполезный и я его закомментила

%Here we evaluated the model for fallings and not fallings separately  (fig. \ref{pic17}). The result on Not Fallings is the best (up to 0.9825).

%\begin{figure}[h!]
    %\centering
    %\includegraphics[scale=0.4]{fallings_and_not.png}
    %\caption{Fallings And Not Fallings}
    %\label{pic17}
%\end{figure}

\newpage
\subsection{Built-in models}

\begin{figure}[h!]
    \centering
    \includegraphics[scale=0.08]{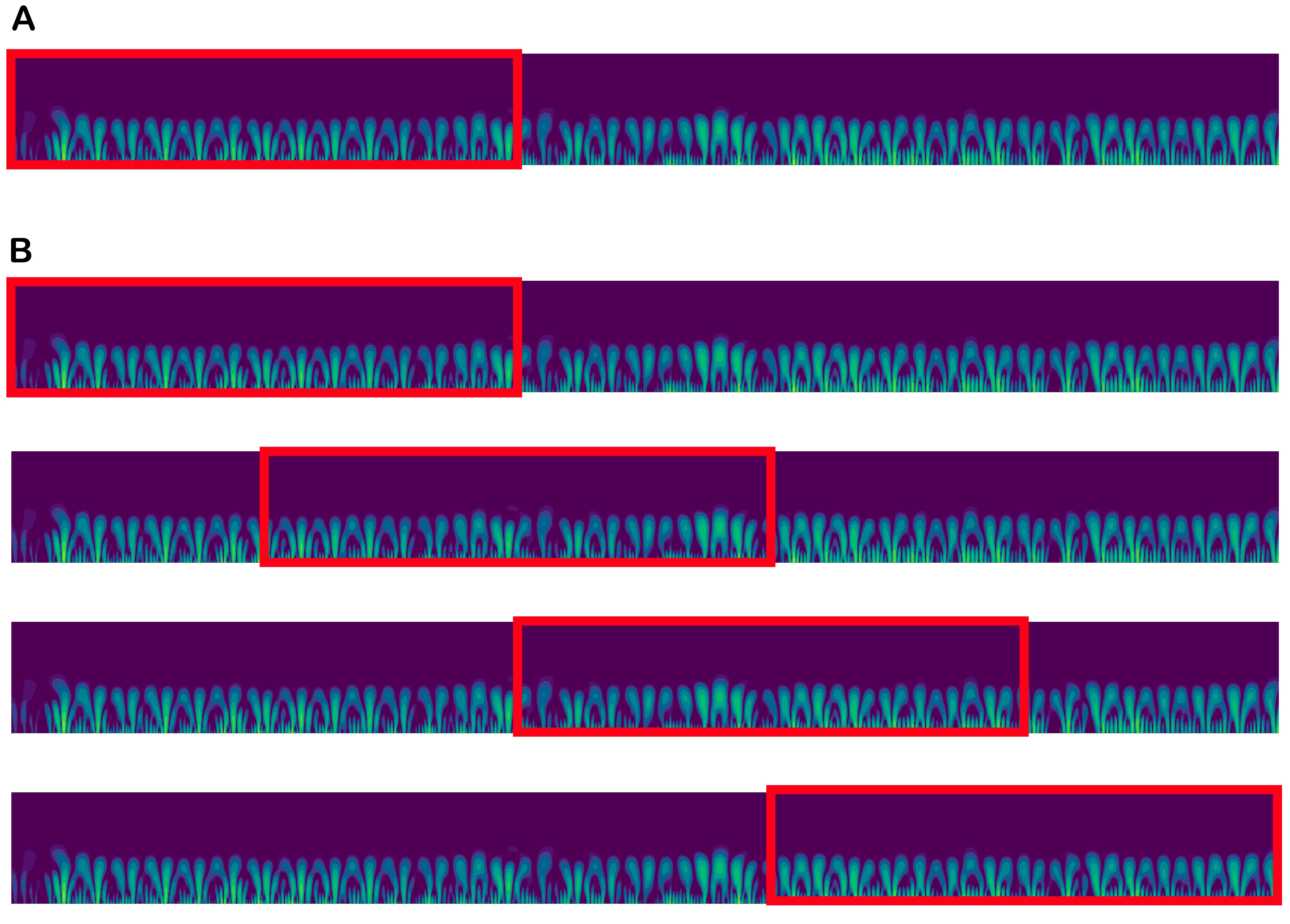}
    \caption{Principle schemes of image cropping. \textit{A} - For CNNs only a single image is obtained from every initial image. \textit{B} - For ResNets a set of images is obtained for every initial image which is achieved by shifting the cropping frame}
    \label{pic17}
\end{figure}

After having researched parameters of our nets up and down, we test the built-in neural networks. We test 6 residual models available in Keras: ResNet50, ResNet101, ResNet152, ResNet50V2, ResNet101V2 and ResNet152V2. For ResNets the accuracy spikes down, so we augment the dataset by cropping every long image not into one short image as before but into a few images with a shift (Fig. \ref{pic17}) and then be increase the batch size up to 500 samples. The best accuracy (99.25 \%) is obtained via implementing the ResNet50 network .

%In order to obtain higher results we augment our dataset by cropping every long image not into one short image as before but into a few images. 

%Finally, we used built-in residual models: ResNet50, ResNet101, ResNet152, ResNet50V2, ResNet101V2 and ResNet152V2 (тут будет краткое описание их структур). It helped to reach the accuracy equal to 0.9834 
\newpage
\begin{figure}[h!]
    \centering
    \includegraphics[scale=0.6]{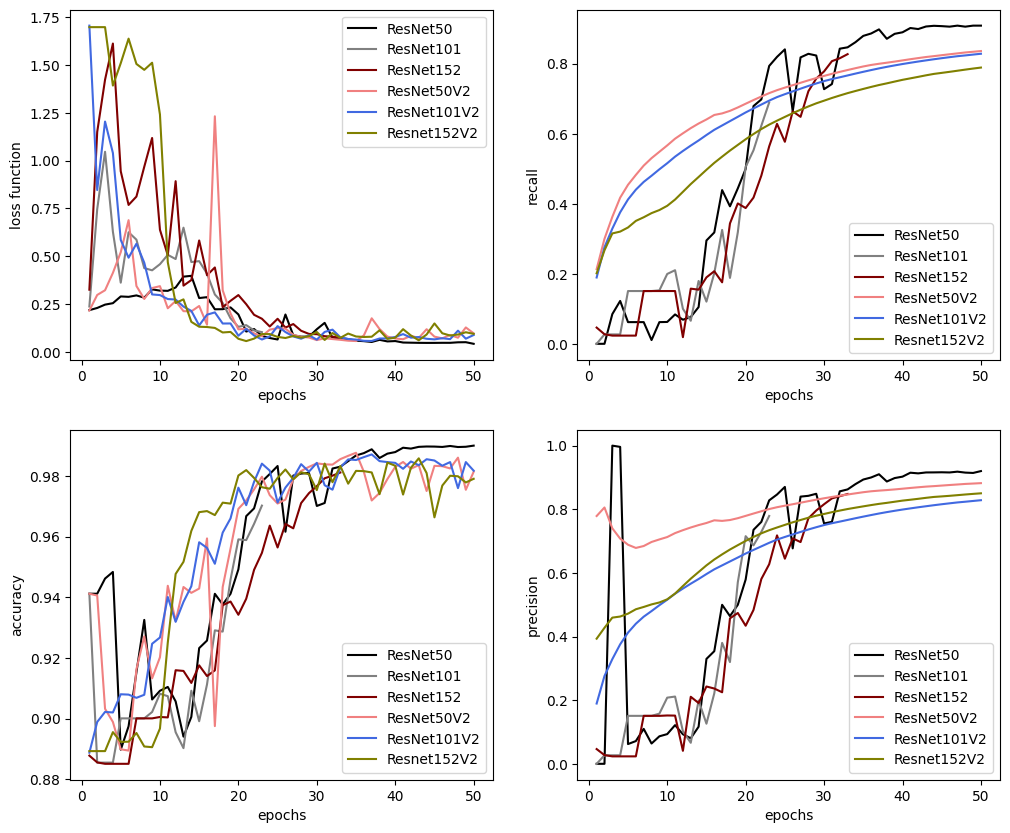}
    \caption{Loss-function, recall, accuracy and precision vs number of epochs for Residual Networks}
    \label{pic18}
\end{figure}

\newpage

\subsection{Image size}

\begin{figure}[h!]
    \centering
    \includegraphics[scale=0.6]{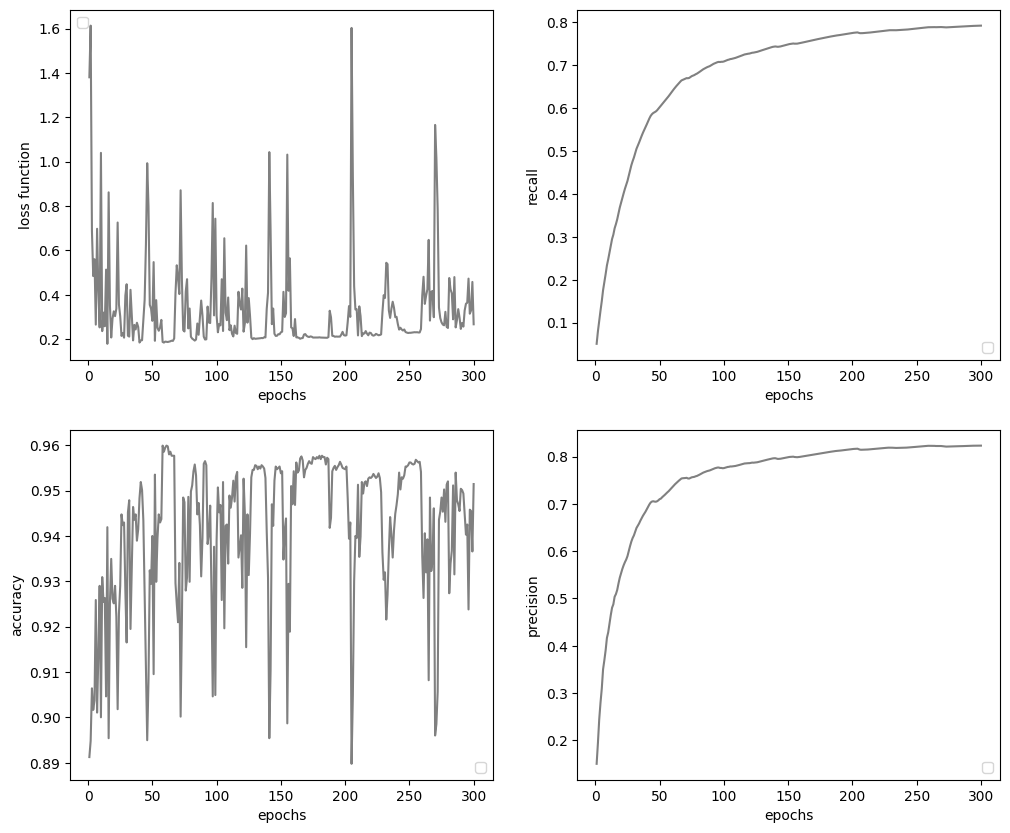}
    \caption{Loss-function, recall, accuracy and precision vs number of epochs for bigger images}
    \label{pic19}
\end{figure}

We also wonder if with a bigger image size the accuracy of prediction will increase. To challenge this theory, we generate images with higher resolution and test them on the ResNet50V2 network. However, the results of prediction become worse (Fig. \ref{pic19}) as while the number of weights increases the number of features remains the same.

\newpage

\subsection{Wavelets with different zero moments}

\begin{figure}[h!]
    \centering
    \includegraphics[scale=0.6]{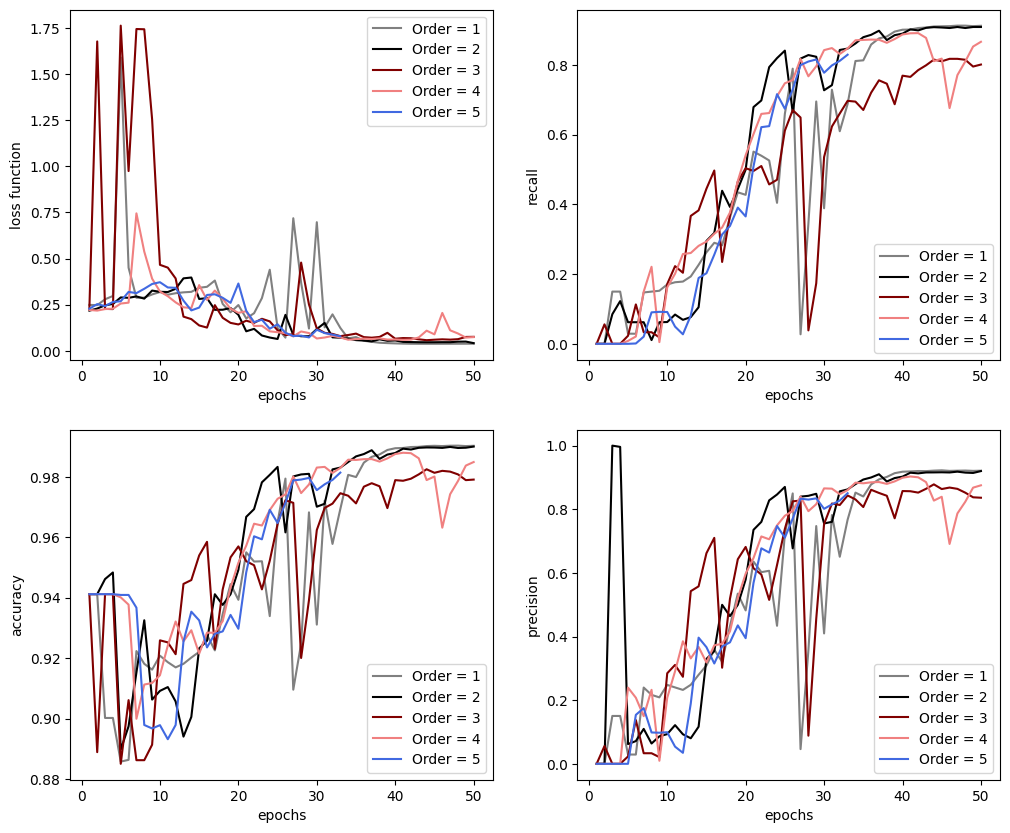}
    \caption{Loss-function, recall, accuracy and precision vs number of epochs for different orders of DOG mother wavelets}
    \label{pic20}
\end{figure}

The order of zero moment of the mother wavelet is also a significant parameter which influences the ability of extracting features. We generate images with 1, 2, 3, 4 and 5 orders of zero moment for the DOG (derivatives of gaussian function) mother wavelets and find out that the best result (Fig. \ref{pic20}) is obtained with the 2nd order (Mexican Hat).

\begin{figure}[h!]
    \centering
    \includegraphics[scale=0.6]{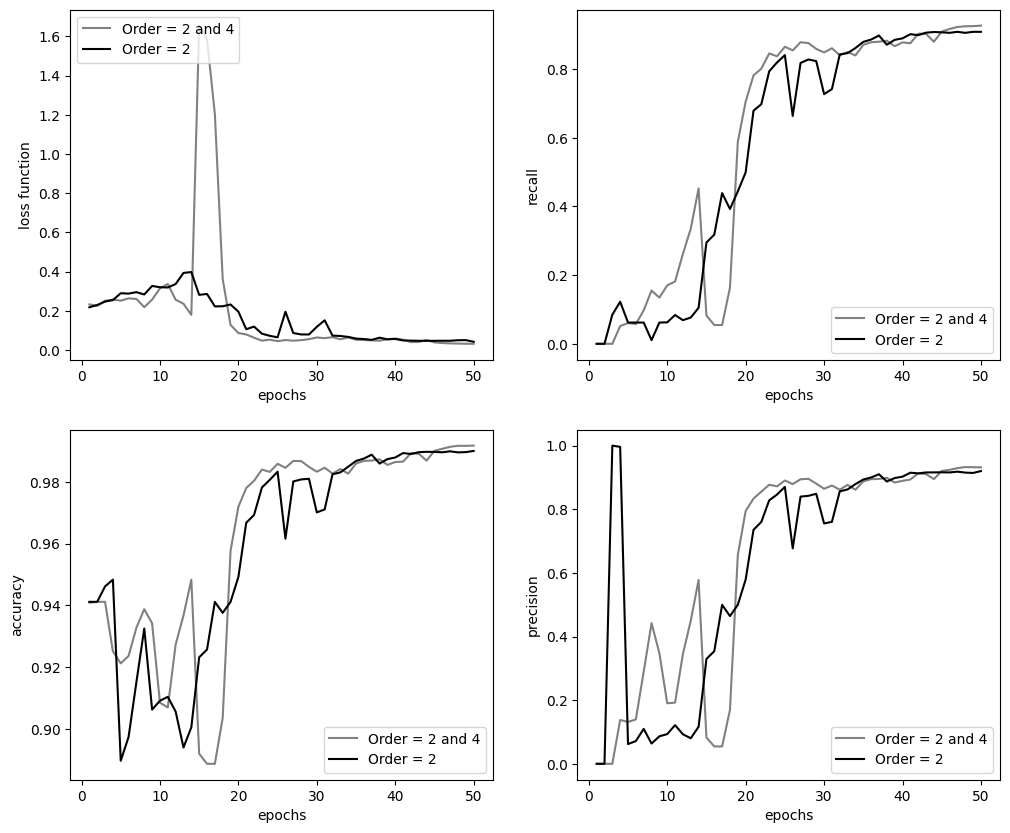}
    \caption{Loss-function, recall, accuracy and precision vs number of epochs for a combination of 2- and 4-orders of mother wavelets}
    \label{pic21}
\end{figure}

\newpage
Finally, we study a combination of images created with 2- and 4-order zero moments (Fig. \ref{pic21}); for this combination the best accuracy is 99.26 \%, thus, a combination of wavelets with different orders of zero moments does not result in a significantly higher accuracy.

\if 0

The following graph(fig. \ref{pic5}) describes how does the number of epochs influence the accuracy of the method. It can be seen that the accuracy estimation carried out without cross-validation increases unlimitedly. At the same time, the accuracy measured via cross-validation has a limitation at about 0.945. This is attributed to the overfitting of the model. As it can be seen from the graph, the most appropriate number of epochs is equal to 8.

\begin{figure}[h!]
    \centering
    \includegraphics[scale=0.45]{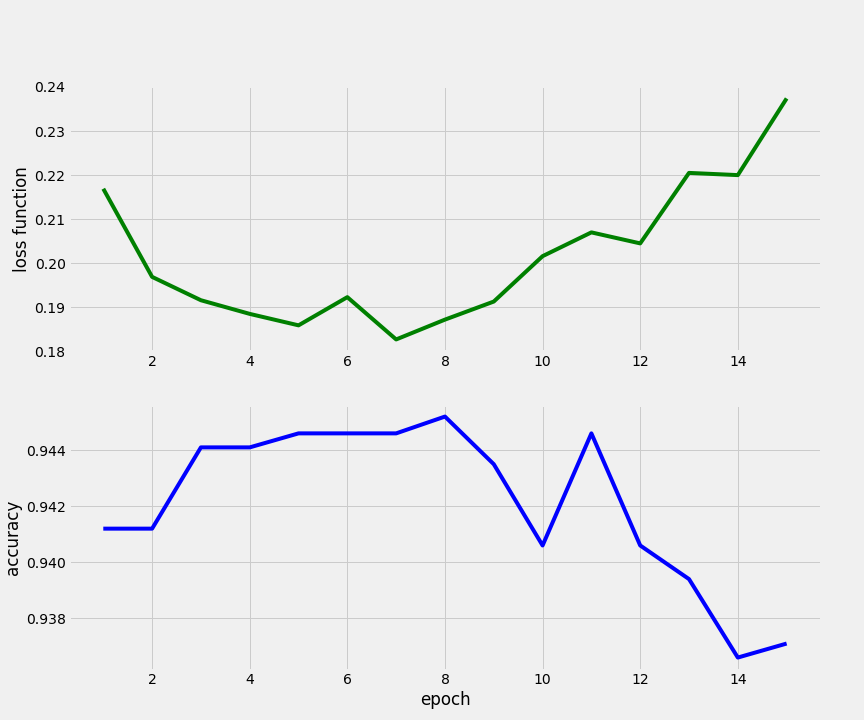}
    \caption{Accuracy and loss-function vs number of epochs}
    \label{pic5}
\end{figure}

\fi

%\newpage

\section{Conclusion}

In the present work we suggest and realize a method for human activity recognition based on continuous wavelet transform and neural networks (code: \textit{https://github.com/AnitaEpsilon/HAR/}). We propose a structure of a convolutional neural network and after varying such parameters as number of convolutional and dense layers, number of neurons in a layer, batchsize, number of spatial axes, mother wavelet, number of mother wavelets etc, we succeed to build an optimal model. We compare the CNN model with ResNets models provided by the \textit{Keras} library. The ResNet models demonstrate a significantly better result. We also figure out, that the $2^{nd}$ order of zero moment is the most optimal parameter for the \textit{DOG} mother wavelet.

Compared with the reference article, we obtain a significantly higher accuracy (99.26\% vs. 93.4\%). 
We consider that the results we received not only play a significant role in the sphere of human activity recognition, but also bring new opportunities in the field of signal classification generally. In the future we plan to continue our research and to develop a device for the real-time activity prediction.

\section{Acknowledgement}

We express our gratitude to Aleksandr Bukharev from Philips Innovation Lab for support, productive discussions and helpful remarks.

%\newpage


\begin{thebibliography}{3}

\bibitem{har}
Vrigkas M, Nikou C and Kakadiaris IA (2015) A Review of Human Activity Recognition Methods. Front. Robot. AI 2:28. doi: 10.3389/frobt.2015.00028

\bibitem{home}
Du Y, Lim Y, Tan Y. A Novel Human Activity Recognition and Prediction in Smart Home Based on Interaction. Sensors. 2019; 19(20):4474.

\bibitem{cognitive}
S. Feng, P. Setoodeh, and S. Haykin, ‘‘Smart home: Cognitive interactive
people-centric Internet of Things,’’ IEEE Commun. Mag., vol. 55, no. 2,
pp. 34–39, Feb. 2017.

\bibitem{areas}
1. Ranasinghe S, Al Machot F, Mayr HC. A review on applications of activity recognition systems with regard to performance and evaluation. International Journal of Distributed Sensor Networks. August 2016. doi:10.1177/1550147716665520

\bibitem{syndrom}
S. Mekruksavanich, N. Hnoohom and A. Jitpattanakul, "Smartwatch-based sitting detection with human activity recognition for office workers syndrome," 2018 International ECTI Northern Section Conference on Electrical, Electronics, Computer and Telecommunications Engineering (ECTI-NCON), Chiang Rai, 2018, pp. 160-164, doi: 10.1109/ECTI-NCON.2018.8378302.

\bibitem{sport}
A. Avci, S. Bosch, M. Marin-Perianu, R. Marin-Perianu and P. Havinga, "Activity Recognition Using Inertial Sensing for Healthcare, Wellbeing and Sports Applications: A Survey," 23th International Conference on Architecture of Computing Systems 2010, Hannover, Germany, 2010, pp. 1-10.
  
\bibitem{wear}
O. D. Lara and M. A. Labrador, "A Survey on Human Activity Recognition using Wearable Sensors," in IEEE Communications Surveys \& Tutorials, vol. 15, no. 3, pp. 1192-1209, Third Quarter 2013, doi: 10.1109/SURV.2012.110112.00192.

\bibitem{UniMiB}
Micucci, D.; Mobilio, M.; Napoletano, P. UniMiB SHAR: A Dataset for Human Activity Recognition Using Acceleration Data from Smartphones. Appl. Sci. 2017, 7, 1101.


%\bibitem{acc_knn}
%P. Maziewski, A. Kupryjanow, K. Kaszuba and A. Czyżewski, "Accelerometer signal pre-processing influence on human activity recognition," Signal Processing Algorithms, Architectures, Arrangements, and Applications SPA 2009, Poznan, 2009, pp. 95-99.

\bibitem{daily_active}
T. Ivascu, K. Cincar, A. Dinis and V. Negru, "Activities of daily living and falls recognition and classification from the wearable sensors data," 2017 E-Health and Bioengineering Conference (EHB), Sinaia, 2017, pp. 627-630, doi: 10.1109/EHB.2017.7995502.

\bibitem{naive}
M. Martinez-Arroyo and L. E. Sucar, "Learning an Optimal Naive Bayes Classifier," 18th International Conference on Pattern Recognition (ICPR'06), Hong Kong, 2006, pp. 1236-1239, doi: 10.1109/ICPR.2006.748.

\bibitem{bayes_ex}
A. M. Jehad Sarkar, Young-Koo Lee \& Sungyoung Lee (2010) A Smoothed Naive Bayes-Based Classifier for Activity Recognition, IETE Technical Review, 27:2, 107-119, DOI: 10.4103/0256-4602.60164


\bibitem{svm}
Terrence S. Furey, Nello Cristianini, Nigel Duffy, David W. Bednarski, Michèl Schummer, David Haussler, Support vector machine classification and validation of cancer tissue samples using microarray expression data , Bioinformatics, Volume 16, Issue 10, October 2000, Pages 906–914, https://doi.org/10.1093/bioinformatics/16.10.906

\bibitem{tree}
S. R. Safavian and D. Landgrebe, "A survey of decision tree classifier methodology," in IEEE Transactions on Systems, Man, and Cybernetics, vol. 21, no. 3, pp. 660-674, May-June 1991, doi: 10.1109/21.97458.


\bibitem{forest}
Combining Deep Learning with Traditional Machine Learning to Improve Classification Accuracy on Small Datasets

\bibitem{knn}
Padraig Cunningham and Sarah Jane Delany, k-Nearest Neighbour Classifiers: 2nd Edition (with Python examples), 2020


\bibitem{knn_ex}
T. Ivascu, K. Cincar, A. Dinis and V. Negru, "Activities of daily living and falls recognition and classification from the wearable sensors data," 2017 E-Health and Bioengineering Conference (EHB), Sinaia, 2017, pp. 627-630, doi: 10.1109/EHB.2017.7995502.



\bibitem{regr}
S. Dreiseitl, L. Ohno-Machado
Logistic regression and artificial neural network classification models: A methodology review
Journal of Biomedical Informatics, 35 (5–6) (2002), pp. 352-359

\bibitem{LR}
Ignatov, A.D., Strijov, V.V. Human activity recognition using quasiperiodic time series collected from a single tri-axial accelerometer. Multimed Tools Appl 75, 7257–7270 (2016). https://doi.org/10.1007/s11042-015-2643-0

\bibitem{ann}
A. K. Jain, Jianchang Mao and K. M. Mohiuddin, "Artificial neural networks: a tutorial," in Computer, vol. 29, no. 3, pp. 31-44, March 1996, doi: 10.1109/2.485891.


\bibitem{yeg}
B. Yegnanarayana
Artificial Neural Networks, Prentice-Hall, New Delhi, India (1999)

\bibitem{four}
Song-Mi Lee, Sang Min Yoon and Heeryon Cho, "Human activity recognition from accelerometer data using Convolutional Neural Network," 2017 IEEE International Conference on Big Data and Smart Computing (BigComp), Jeju, 2017, pp. 131-134, doi: 10.1109/BIGCOMP.2017.7881728.



\bibitem{wu}
J. Wu
Introduction to convolutional neural networks
Natl. Key Lab Novel Softw. Technol., Nanjing, China,
Nov. 2016, pp. 1–28.

\bibitem{ensemble}
Mukherjee, D., Mondal, R., Singh, P.K. et al. EnsemConvNet: a deep learning approach for human activity recognition using smartphone sensors for healthcare applications. Multimed Tools Appl (2020). https://doi.org/10.1007/s11042-020-09537-7



\bibitem{modified}
A. Keshavarzian, S. Sharifian and S. Seyedin, "Modified deep residual network architecture deployed on serverless framework of iot platform based on human activity recognition application", Future Generation Computer Systems, vol. 101, pp. 14-28, 2019.

\bibitem{ronao}
C.A. Ronao, S.-B. Cho
Human activity recognition with smartphone sensors using deep learning neural networks
Expert Syst. Appl., 59 (2016), pp. 235-244

\bibitem{neurohive}
ResNet (34, 50, 101): Residual CNNs for Image Classification Tasks,
23 January 2019; Available: 
https://neurohive.io/en/popular-networks/resnet/


\bibitem{residual}
Deep Residual Learning for Image Recognition;
Kaiming He Xiangyu Zhang Shaoqing Ren Jian Sun;
Microsoft Research


\bibitem{mit_news}
Artificial intelligence model detects asymptomatic Covid-19 infections through cellphone-recorded coughs,
Jennifer Chu | MIT News Office,
Publication Date:October 29, 2020

\bibitem{fingerprints} 
Nahar, S. Tanwani and N. Chaudhari, "Fingerprint classification using deep neural network model resnet50", IJRAR, vol. 5, pp. 4, December 2018.

\bibitem{transport}
Sebastien Richoz, Andres Perez-Uribe, Philip Birch, and Daniel Roggen. 2019. Benchmarking deep classifiers on mobile devices for vision-based transportation recognition. In Adjunct Proceedings of the 2019 ACM International Joint Conference on Pervasive and Ubiquitous Computing and Proceedings of the 2019 ACM International Symposium on Wearable Computers (UbiComp/ISWC '19 Adjunct). Association for Computing Machinery, New York, NY, USA, 803–807. DOI:https://doi.org/10.1145/3341162.3344849


%\bibliographystyle{gost780s} % ГОСТ 7.80
%\bibliography{ref} % MachLearn.bib


%\bibitem{convert}
%An efficient way to convert 1D signal to 2D digital image using energy values Странное цитирование

\bibitem{abel}
The Abel Prize Laureate 2017;
Available: https://www.abelprize.no/c69461/binfil/download.php?tid=69541

\bibitem{comparison}
Application of Wavelet Transform and its Advantages
Compared to Fourier Transform
M. Sifuzzaman1
, M.R. Islam1
 and M.Z. Ali;
 available: http://inet.vidyasagar.ac.in:8080/jspui/handle/123456789/779

\bibitem{records}
Peng, Pingan, et al. "Automatic classification of microseismic records in underground mining: a deep learning approach." IEEE Access 8 (2020): 17863-17876.

\bibitem{kr}
Крючкова И. В. Ряды и преобразование Фурье: методические указания. – 2011. Available: http://elib.osu.ru/handle/123456789/8947

\bibitem{from}
Gao R and Yan R 2011 From Fourier Transform to Wavelet Transform: A Historical
Perspective Springer Wavelets 17-32

\bibitem{short}
S. Nawab, T. Quatieri and Jae Lim, "Signal reconstruction from short-time Fourier transform magnitude," in IEEE Transactions on Acoustics, Speech, and Signal Processing, vol. 31, no. 4, pp. 986-998, August 1983, doi: 10.1109/TASSP.1983.1164162.

\bibitem{uncertainty}
L. Cohen, "The uncertainty principle in signal analysis," Proceedings of IEEE-SP International Symposium on Time- Frequency and Time-Scale Analysis, Philadelphia, PA, USA, 1994, pp. 182-185, doi: 10.1109/TFSA.1994.467263.

\bibitem{har_short}
S. Matsui, N. Inoue, Y. Akagi, G. Nagino and K. Shinoda, "User adaptation of convolutional neural network for human activity recognition," 2017 25th European Signal Processing Conference (EUSIPCO), Kos, 2017, pp. 753-757, doi: 10.23919/EUSIPCO.2017.8081308.

\bibitem{polikar}
R. Polikar, The Wavelet Tutorial. [Online]. Available: http://users.rowan.edu/~polikar/WTtutorial.html

\bibitem{marry}
R. J. E. Merry, Wavelet Theory and Application: A Literature Study,
DCT 2005.53. Eindhoven, The Netherlands: Eindhoven Univ.
Technol., 2005.

\bibitem{fast}
O. Rioul and P. Duhamel, "Fast algorithms for discrete and continuous wavelet transforms," in IEEE Transactions on Information Theory, vol. 38, no. 2, pp. 569-586, March 1992, doi: 10.1109/18.119724.
Copy

\bibitem{sad}
J Sadowsky, Investigation of signal characteristics using the continuous wavelet transform
johns hopkins apl technical digest, 1996 - jhuapl.edu
Neural Computation
Volume 29 | Issue 9 | September 2017
p.2352-2449

\bibitem{yakovlev}
Яковлев А. Н. Введение в вейвлет-пpеобpазования: учеб.
пособие. Новосибиpск: Изд-во НГТУ, 2003. 104 с.

\bibitem{zero}
C. S. Burrus and J. E. Odegard, "Wavelet systems with zero moments," Proceedings of the 1998 IEEE International Conference on Acoustics, Speech and Signal Processing, ICASSP '98 (Cat. No.98CH36181), Seattle, WA, USA, 1998, pp. 1541-1544 vol.3, doi: 10.1109/ICASSP.1998.681744.

\bibitem{automatic}
Rajeul, S. "Automatic microseismic signals classification with Deep Learning using multi-input Convolutional Neural Networks." Second EAGE Workshop on Machine Learning. Vol. 2021. No. 1. European Association of Geoscientists \& Engineers, 2021.

\bibitem{cardio}
Byeon, Yeong-Hyeon, Sung-Bum Pan, and Keun-Chang Kwak. "Intelligent deep models based on scalograms of electrocardiogram signals for biometrics." Sensors 19.4 (2019): 935.

\bibitem{dcnn}
Waseem Rawat and Zenghui Wang;
Deep Convolutional Neural Networks for Image Classification: A Comprehensive Review
Posted Online August 23, 2017
© 2017 Massachusetts Institute of Technology
 %https://www.mitpressjournals.org/doi/full/10.1162/neco_a_00990 )
 
\bibitem{He_k}
Kaiming He, Xiangyu Zhang, Shaoqing Ren, Jian Sun; Proceedings of the IEEE Conference on Computer Vision and Pattern Recognition (CVPR), 2016, pp. 770-778

\bibitem{earth}
Chen, Yangkang, et al. "Automatic waveform classification and arrival picking based on convolutional neural network." Earth and Space Science 6.7 (2019): 1244-1261.


\end{thebibliography}
\end{document}